%% file: main.tex
\documentclass[sigconf,preprint]{acmart}

\AtBeginDocument{%
  }

\setcopyright{acmlicensed}
\copyrightyear{2027}
\acmYear{2027}
\acmDOI{XXXXXXX.XXXXXXX}
\acmConference[KDD '27]{33rd ACM SIGKDD Conference on Knowledge Discovery and Data Mining}{August 01--05,
  2027}{San Jose, CA}
\acmISBN{978-1-4503-XXXX-X/2027/08}

\usepackage{xspace}

\theoremstyle{remark}
\newtheorem{definition}{Definition}

\newtheoremstyle{problemstyle}  % <name>
        {3pt}                                               % <space above>
        {3pt}                                               % <space below>
        {\normalfont}                               % <body font>
        {1em}                                                  % <indent amount}
        {\bfseries\itshape}                 % <theorem head font>
        {\normalfont\bfseries:}         % <punctuation after theorem head>
        {.5em}                                          % <space after theorem head>
        {}                                                  % <theorem head spec (can be left empty, meaning `normal')>
\theoremstyle{problemstyle}

\usepackage{makecell}
\usepackage{tabularx}
\usepackage[linesnumbered,ruled,vlined]{algorithm2e}
\usepackage{bbm}
\usepackage{amsmath}
\usepackage{scalerel}
\usepackage{tikz}
\usepackage[inline]{enumitem}
\usetikzlibrary{plotmarks}
\usetikzlibrary{patterns}
\usepackage{pgfplots}
\usepackage{pgfplotstable}
\usetikzlibrary{pgfplots.statistics}
\usepackage{subfigure}
\usepackage{graphicx}
\usepackage{fontawesome5}
\usepackage{cleveref}
\usepackage{setspace}
\usepackage{enumitem}
\usepackage{balance}
\usepackage{url}
\usepackage{fvextra}
\usepackage{multirow}
\usepackage{multicol}
\usepackage{algorithmic}
\usepackage{pifont}    % for ding symbols
\usepackage[inline]{enumitem}
\newcommand{\method}{FormStruct-Bench\xspace}

\newcommand{\hflogo}{%
  \raisebox{-0.15\height}{\includegraphics[height=1.05em]{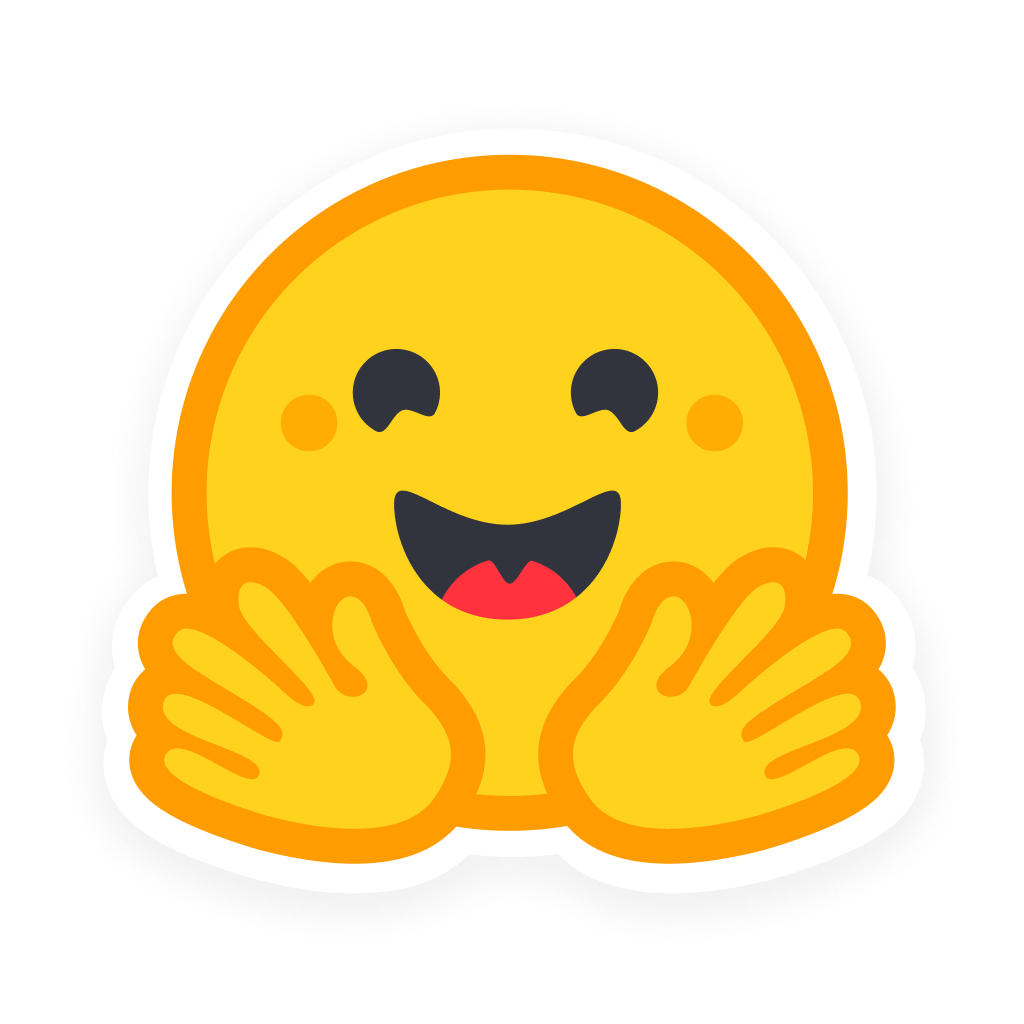}}%
}
\definecolor{githubpurple}{HTML}{6E5494}
\newcommand{\githublogo}{\textcolor{githubpurple}{\faGithub}}
\newcommand{\circledchar}[2][gray!70]{%
    \tikz[baseline=(char.base)]{
        \node[shape=circle,draw=#1,fill=#1,text=white,inner sep=0.75pt,scale=0.85] (char) {#2};
    }
    }
\begin{document}

\title{\method: A Hierarchical and Diagnostic Benchmark for Table-Form Document Structure Recognition}

\author{Lujie Ban$^{1}$, Jiangtao Zhu$^{1}$, Yuanheng Yu$^{2}$, Jiasheng Shi$^{1}$, Chenhao Ma$^{*1}$}
\affiliation{%
  \institution{$^{1}$ The Chinese University of Hong Kong, Shenzhen,$^{2}$ University of Washington}
  \city{Shenzhen}
  \country{China}}
\email{\{lujieban,jiangtaozhu\}@link.cuhk.edu.cn, yuanhy@uw.edu, \{shijiasheng,machenhao\}@cuhk.edu.cn}
% \author{Jiasheng Shi}
% \affiliation{%
%   \institution{School of Data Science, The Chinese University of Hong Kong, Shenzhen}
%   \city{Shenzhen}
%   \country{China}}
% \email{shijiasheng@cuhk.edu.cn}

% \author{Chenhao Ma$^*$}
% \affiliation{%
%   \institution{School of Data Science, The Chinese University of Hong Kong, Shenzhen}
%   \city{Shenzhen}
%   \country{China}}
% \email{machenhao@cuhk.edu.cn}

\renewcommand{\shortauthors}{Ban et al.}

\begin{abstract}
\input{sections/abstract}
\end{abstract}




\received{20 February 2007}
\received[revised]{12 March 2009}
\received[accepted]{5 June 2009}

\maketitle

\input{sections/introduction}
\input{sections/methodology}
\input{sections/data_statistics}

\input{sections/evaluation}
\input{sections/experiment}
\input{sections/related_work}
\input{sections/conclusion}
\newpage
\bibliographystyle{ACM-Reference-Format}
\bibliography{main}

\clearpage
\appendix
\input{sections/appendix}

\end{document}

%% file: sections/abstract.tex
Transforming table-form documents into machine-processable records requires recovering not only their visible content but also the multilevel structure that organizes it.
However, existing benchmarks evaluate either holistic document outputs or conventional table grids, and their aggregate scores provide little insight into where structural failures occur.
We introduce \method, a hierarchical and diagnostic benchmark that evaluates table-form document structure recognition at both the document level and progressively finer component levels, allowing aggregate performance to be traced back to specific structural failure modes.
To construct auditable ground truth at scale, we annotate 70 reusable templates and expand them into 7,000 verified instances through a provenance-preserving Director--Artist--Verifier pipeline; all 1,100 instances in the template-disjoint test set additionally receive human review.
Our evaluation protocol uses five primary metrics and three structure-specific diagnostics across page, schema, and component levels, together with slices over difficulty, structural constraints, and visual degradation.
Across 14 API-hosted and locally deployable systems plus two SFT variants, the best document-level score reaches 83.85\%, whereas the best reported fine-grained structural score remains below 18\%.
These results reveal a pronounced gap between reading document content and recovering the hierarchy and regional organization required for reliable table-form understanding.

%% file: sections/introduction.tex
\section{Introduction}

% Tables are a fundamental structured representation in document understanding: they organize values, attributes, and constraints through spatial layouts that downstream systems can query, validate, and transform~\cite{zhang2020webtable,yin2020tabert,deng2020turl,tablebank2020,pubtables1m2021}. A practically important subclass is \emph{form-like tables}, where tabular organization is used not only to present records, but also to collect, verify, and relate user-provided information~\cite{funsd2019,aggarwal2021multimodal,docile2023}. They store a wealth of information essential for knowledge discovery~\cite{cafarella2008webtables,zhang2020webtable}, decision support~\cite{shim2002past}, business analytics~\cite{chen2012business}, compliance auditing~\cite{appelbaum2017big}, and data-driven document intelligence~\cite{layoutlm2020,appalaraju2021docformer}. In these applications, the central requirement is not merely to read visible text, but to recover the structural information into processable records. This requirement is the core of table structure recognition (TSR)~\cite{scitsr2019,pubtabnet2020,pubtables1m2021}, which converts table images into explicit structural representations such as HTML or JSON, as shown in \Cref{fig:tsr_def}.
% \begin{figure}[ht]
%     \centering
%     \includegraphics[width=\linewidth]{figures/problem_definition.pdf}
%     \caption{Form-like table structure recognition (TSR) overview.}
%         \label{fig:tsr_def}
% \end{figure}

\begin{figure}[ht]
  \centering
  \includegraphics[width=0.9\linewidth]{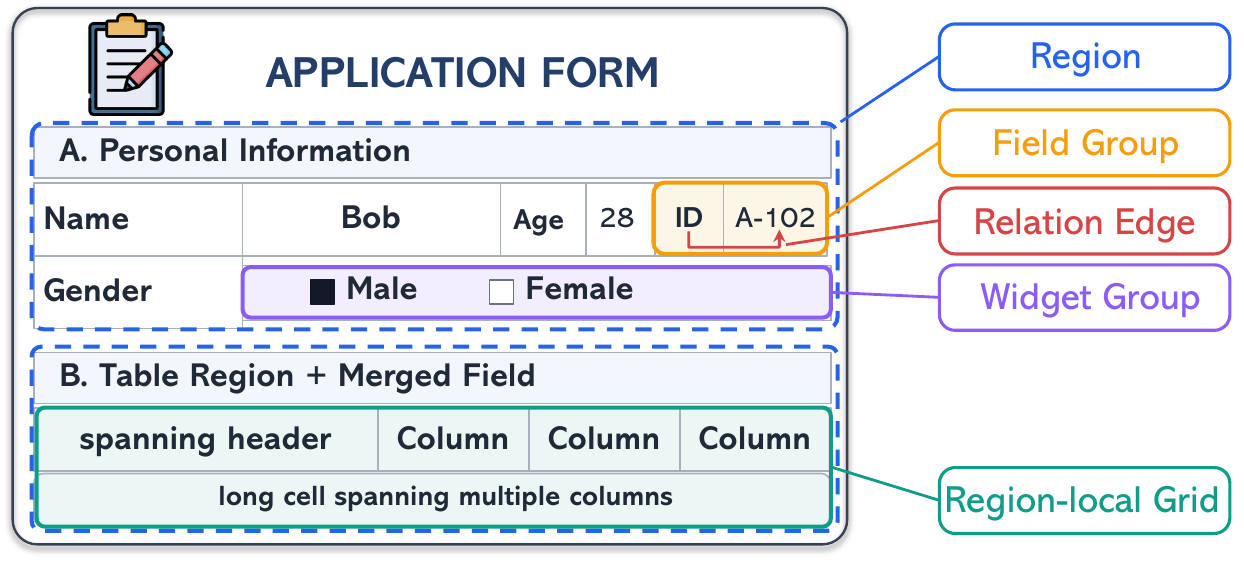}
  \caption{An example table-form document with its structural annotation: semantic regions, a region-local grid, field and widget groups, and relation edges.}
  \label{fig:problem_def}
\end{figure}

Documents are a primary carrier of real-world information and an essential source of data for automated information processing~\cite{layoutlm2020,appalaraju2021docformer}. 
Among them, forms constitute a particularly important document class, providing standardized interfaces through which organizations collect, validate, and exchange information~\cite{funsd2019,docile2023}. Table-form documents, such as application, tax, and medical-intake forms, organize such information through table-like layouts and are central to record ingestion~\cite{sroie2021,docile2023}, compliance auditing~\cite{appelbaum2017big}, and automated workflow execution~\cite{vanDerAalst2018rpa,chakraborti2020ipa}.
Their information is encoded not only in visible text but also in structure: a page is divided into semantic sections, a section may carry its own local grid, and checkboxes and other widgets take their meaning from the fields and labels they are linked to (\Cref{fig:problem_def})~\cite{aggarwal2021multimodal,lee2022formnet}.
Consequently, transforming a table-form document into reliable, machine-processable records requires recovering these structures rather than merely reading the visible text.

This transformation is a central goal of document understanding~\cite{layoutlm2020,appalaraju2021docformer}.
With the rise of LLMs, document understanding has advanced rapidly from localized text and field extraction toward end-to-end document parsing and structured generation, with models jointly reasoning over visual, textual, and layout signals~\cite{layoutlmv32022,donut2022,image2struct2024,receiptbench2026}.
In parallel, Table Structure Recognition (TSR) has focused on the precise reconstruction of conventional table layouts, including row--column topology, cell spans, and explicit grid or markup representations~\cite{scitsr2019,pubtabnet2020,pubtables1m2021,grits2023}.

However, table-form documents fall between these two lines of research. Document understanding extracts fields and answers questions over forms but rarely requires faithful reconstruction of the underlying structure, whereas TSR reconstructs structure precisely but confines its output to a single row--column grid, leaving fields, widgets, and their semantic relations outside its scope. Recognizing table-form documents demands both: the geometric precision of TSR, applied to the hierarchical and relational organization studied in document understanding. We term this problem table-form document structure recognition (\Cref{fig:problem_def}).
This broader recognition target also creates a distinct evaluation challenge. 
Existing benchmarks typically evaluate either document-level understanding through question answering, information extraction, and holistic document parsing~\cite{docvqa2021,docile2023,omnidocbench2025}, or conventional table reconstruction through canonical grid and markup recovery~\cite{scitsr2019,pubtabnet2020,pubtables1m2021,grits2023}, but neither benchmark family fully captures the joint structural requirements of table-form documents, as summarized in \Cref{tab:benchmark_comparison}.
%
% It therefore remains unclear whether current benchmarks provide an adequate basis for evaluating table-form document structure recognition.
Current benchmarks therefore do not provide an adequate basis for evaluating table-form document structure recognition.

More specifically, the missing component is diagnostic evaluation: current benchmarks often report whether a predicted structure matches the target, but not the root cause of the failure.
For example, a visa or medical form may place personal information, eligibility questions, checkboxes, and signature fields in different regions of the same page; the value of one field is often determined not only by nearby text, but also by its section, option group, and relation to distant labels.
%
% Form-oriented and general structure-extraction benchmarks have expanded evaluation beyond conventional tables, but they still tend to emphasize extracted fields or flexible output formats rather than diagnosing whether a model has recovered the structure that makes a form-like table usable~\cite{funsd2019,docile2023,receiptbench2026,image2struct2024}.
%
Therefore, current evaluations make it difficult to answer a question \circledchar{Q}: when a model fails on a table-form document, which part of the structure caused the failure? Answering this question requires fine-grained structural annotations, which are difficult to obtain reliably from automatic LLM labeling, while instance-level human annotation is difficult to scale.

\begin{table}[t]
\centering
\caption{Comparison of structural targets and diagnostic capabilities.
$\triangle$ denotes partial or indirect support.}
\label{tab:benchmark_comparison}
\scriptsize
\setlength{\tabcolsep}{2.0pt}
\renewcommand{\arraystretch}{0.92}
\begin{tabular}{lcccccc}
\toprule
Benchmark &
\makecell{Grid\\structure} &
\makecell{Full-page\\layout} &
\makecell{Form\\elements} &
\makecell{Structural\\relations} &
\makecell{Component\\metrics} &
\makecell{Diagnostic\\slices} \\
\midrule
SciTSR~\cite{scitsr2019}
  & \checkmark & $\times$ & $\times$ & $\times$ & $\times$ & $\times$ \\
PubTabNet~\cite{pubtabnet2020}
  & \checkmark & $\times$ & $\times$ & $\times$ & $\times$ & $\times$ \\
PubTables-1M~\cite{pubtables1m2021}
  & \checkmark & $\triangle$ & $\times$ & $\times$ & $\triangle$ & $\times$ \\
FUNSD~\cite{funsd2019}
  & $\times$ & \checkmark & \checkmark & \checkmark & $\triangle$ & $\times$ \\
DocILE~\cite{docile2023}
  & $\triangle$ & \checkmark & \checkmark & $\triangle$ & $\triangle$ & $\times$ \\
SRFUND~\cite{ma2024srfund}
  & $\triangle$ & \checkmark & \checkmark & \checkmark & \checkmark & $\times$ \\
Image2Struct~\cite{image2struct2024}
  & $\triangle$ & \checkmark & $\triangle$ & $\triangle$ & $\times$ & $\times$ \\
OmniDocBench~\cite{omnidocbench2025}
  & \checkmark & \checkmark & $\times$ & $\times$ & $\triangle$ & $\triangle$ \\
VAREX~\cite{barzelay2026varex}
  & $\times$ & $\times$ & \checkmark & $\times$ & \checkmark & \checkmark \\
\midrule
\textbf{\method}
  & \checkmark & \checkmark & \checkmark & \checkmark & \checkmark & \checkmark \\
\bottomrule
\end{tabular}
\end{table}
To answer question \circledchar{Q}, we introduce \method, a hierarchical and diagnostic benchmark for table-form document structure recognition.
The key idea is to evaluate table-form documents as structured pages rather than as a single flattened table or a bag of extracted fields.
Each sample is represented through a hierarchy of page regions, region-local grids, field and widget groups, and relation edges (\Cref{fig:problem_def}), which allows evaluation to isolate different structural failure modes.
This representation also raises a data-construction requirement: the labels must remain consistent, traceable, and verifiable across document instances.
To make the labels auditable, \method decouples annotation from generation: human annotators construct reusable structural templates, and a multi-agent generation pipeline expands each template into verified instances while preserving provenance for the source template, constraints, visual parameters, and verification outcomes. This design couples the reliability of human annotation with the scalability of automatic generation.

The same hierarchical representation also defines the evaluation protocol. Instead of reducing each prediction to a single table-level similarity score, \method assesses whether a model recovers each major structural component, scoring predictions at the document level and at progressively finer component levels so that an overall score can be traced back to specific structural failures. It then stratifies these component scores by difficulty, structural constraints, and visual perturbations to expose where each model’s capability boundary lies.

Our contributions are summarized as follows:
\begin{itemize}[leftmargin=*]
\item We construct \method, a benchmark for table-form document structure recognition, using 70 reusable human-annotated templates and 7,000 verified generated instances with auditable provenance.
\item We provide a diagnostic evaluation protocol that combines document-level scoring with finer component-level analysis.
\item We benchmark 14 API-hosted and locally deployable systems plus two SFT variants, revealing a substantial gap between content recovery and structural recognition: the best document-level score exceeds 80\%, while all reported fine-grained structural scores remain below 18\%, with further degradation on relation-dense, grid-heavy, and visually corrupted cases.
\end{itemize}

%% file: sections/methodology.tex
\section{Dataset Construction}
\label{sec:data_construction}
\begin{figure*}[!htp]
    \centering
    \includegraphics[width=\linewidth]{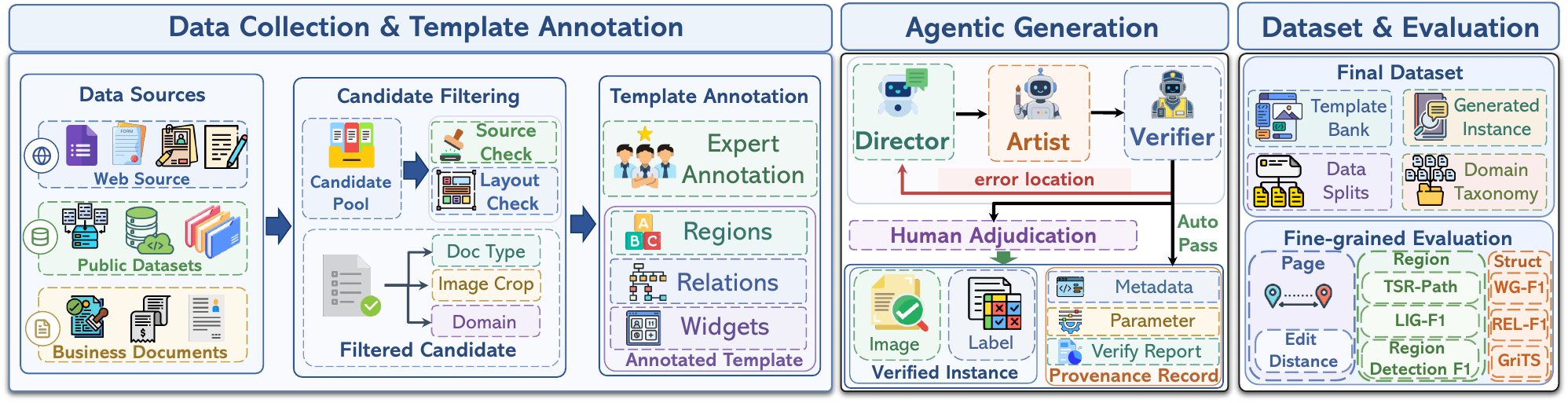}
    \caption{Overview of the \method dataset construction pipeline.}
    \label{fig:framework}
\end{figure*}

We construct \method by converting human-annotated table-form documents into reusable structural templates and expanding each template through a Director--Artist--Verifier pipeline (\Cref{fig:framework}). This design scales instance generation while preserving exact labels, diagnostic constraints, and provenance.

\subsection{Table-Form Document Structure Recognition}
We define the prediction target of table-form document structure recognition as follows.
\begin{definition}[Table-Form Document Structure Recognition]
Given a table-form document image $I$, the task is to predict
\begin{equation}
    \hat{\mathcal{S}} = f(I), \qquad
\mathcal{S} = \big(\mathcal{R},\, \{\mathcal{G}_r\}_{r\in\mathcal{R}},\, \{\mathcal{F}_r\}_{r\in\mathcal{R}},\, \{\mathcal{W}_f\}_{f\in\mathcal{F}},\, \mathcal{E}\big),
\end{equation}
where $\mathcal{R}$ denotes typed semantic regions, $\mathcal{G}_r$ the (possibly empty) local grid of region $r\in\mathcal{R}$, $\mathcal{F}_r$ the fillable fields of region $r$ with $\mathcal{F}=\bigcup_{r\in\mathcal{R}}\mathcal{F}_r$, $\mathcal{W}_f$ the widget groups attached to field $f\in\mathcal{F}$, and $\mathcal{E}$ directed, typed relations among these elements (e.g., key-to-field links).
\end{definition}
This nesting makes $\mathcal{S}$ a hierarchical page structure rather than the single canonical grid of conventional TSR: regions carry local grids and fields, and fields carry widget groups. The evaluation protocol in \Cref{sec:evaluation} follows the same hierarchy: local grids are matched only under paired parent regions, and widget groups only under matching parent fields.

\subsection{Template Collection and Annotation}
Form-like tables combine local grids, fillable fields, selection controls, signature areas, and relations across non-adjacent regions. Because labeling every filled instance is costly, we instead collect reusable templates from web sources and 2 datasets including XFUND \cite{xu-etal-2022-xfund} and Common Crawl \cite{commoncrawl2026}. We retain candidates that combine table-like spatial organization with form-specific elements and whose stable layouts annotators can recover unambiguously; filled values are removed without altering the layout.

Each document is rasterized at normalized resolution and orientation, then annotated as a structural template. The annotation captures table boundaries, cells and spans, visible text, semantic regions, region-local grids, fillable fields, widgets, and structural relations. Rather than imposing a page-wide grid, annotators mark grids locally and represent checkboxes, radio buttons, character boxes, blank lines, and signature areas as fields or widgets. They record observable relations, including key-to-field links, field-to-widget links, section membership, and line-item grouping; constraint tags, counts, and difficulty levels are instead derived from the exported metadata.

\subsection{Multi-agent Instance Generation}
The three agents separate content planning, rendering, and quality control. The Director samples field values, layout parameters, and visual degradation, and assigns each instance a target constraint profile. It can therefore populate specific diagnostic slices deliberately; the difficulty tier is derived from the assigned structural and context constraints rather than labeled post hoc. The resulting metadata fully specifies the rendering job.

Using only this metadata, the Artist renders the image without access to the original document. Geometry and annotations are computed at render time, yielding exact ground truth for typed regions, region-local grid topology, fields, and widget states. The Verifier uses few-shot in-context learning to instruct an LLM to infer the corresponding structured representation from the rendered image. The inferred result is subsequently validated against the reference answer, with any detected inconsistency returned to the Artist for iterative refinement and re-rendering.

\subsection{Verification and Provenance}
Automated verification checks four aspects: whether region types, local grids, and widget states match the metadata; whether OCR recovers the assigned values from fillable regions; whether border visibility, contrast, and skew remain within their assigned ranges; and whether annotations have valid coordinates, field types, and relation targets. Failed samples trigger a structured report and are re-generated until they pass or exhaust the retry budget, after which they are discarded and logged.

Every test sample that passes these checks additionally receives human review to ensure that its regions, relations, and widget states are unambiguously supported by visible evidence. Reviewers mark a sample as accepted, corrected, or discarded; corrected samples are re-verified, and unresolved samples are excluded.

Passing samples receive a provenance record containing the source template, generation metadata, constraint tags, and, for test samples, the review outcome. Instances retain their template's structural skeleton while varying filled values and visual parameters. Consequently, the test set can be partitioned by recorded constraint tags without further annotation, enabling constraint-level diagnostic evaluation.

%% file: sections/data_statistics.tex
\section{Data Statistics}
\label{sec:data_statistics}
Reporting only sample counts and coarse category distributions tells a researcher little about whether a benchmark will expose the failure modes they care about. \method organizes its statistics around three questions: \textbf{Q1.} what structural configurations the templates cover; \textbf{Q2.} which constraint combinations appear in the generated instances; and \textbf{Q3.} whether the test split samples those combinations in sufficient numbers to support per-constraint diagnostic evaluation. Our dataset is publicly available at \hflogo\,\url{https://huggingface.co/datasets/D2I-CUHK-Shenzhen/FormStruct-Bench}.

\subsection{Template-level Statistics}
\label{sec:template_stats}

\begin{figure}[ht]
\centering
\includegraphics[width=\linewidth]{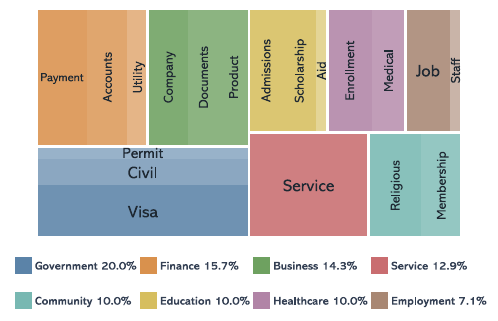}
\caption{Template distribution across coarse and fine domains.}
\label{fig:template_dist}
\end{figure}

\textbf{Q1} concerns domain breadth and structural complexity. Figure~\ref{fig:template_dist} shows that the 70 templates span eight named application domains, one residual unclassified template, and 20 fine-grained labels. Government and immigration is the largest coarse domain (14 templates, 20.0\%), while no fine-grained use case exceeds 9 templates (12.9\%). The calibrated difficulty distribution of 11/24/24/11 templates across L1--L4 further balances routine and challenging structures, reducing dependence on any single domain or layout convention.

Table~\ref{tab:template_stats} shows that templates contain 1--7 semantic sections and 1--9 table regions; the median template has 4 semantic sections, 4 table regions, 23.5 label-field units, 16 selection fields, and 79 structural relation edges. Thus, \method targets page-level form layouts whose local grids, fillable regions, selection controls, and cross-region relations jointly define the recognition target, rather than isolated data tables.

\begin{table}[ht]
\centering
\caption{Template-level structural statistics.}
\label{tab:template_stats}
  \begin{tabular}{lcccc}
  \toprule
  Feature & Min & Median & Mean & Max\\
  \midrule
Semantic sections
  & 1 & 4 & 3.9 & 7 \\
Table regions
  & 1 & 4 & 4.6 & 9 \\
Field groups
  & 0 & 4 & 5.2 & 25 \\
Label-field units
  & 2 & 23.5 & 25.0 & 59 \\
Multi-field units
  & 0 & 4 & 5.7 & 23 \\
Selection fields
  & 0 & 16 & 19.0 & 69 \\
Structural relation edges
  & 38 & 79 & 88.1 & 204 \\
Maximum hierarchy depth
  & 3 & 5 & 4.7 & 8 \\
  \bottomrule
  \end{tabular}
\end{table}

Table~\ref{tab:lang_stats} covers seven languages plus bilingual Chinese--English templates across three script families. Japanese, English, and Chinese account for 74.3\% of the pool; Arabic contributes right-to-left layouts, while German, Spanish, and Portuguese broaden the Latin-script coverage. This mix helps separate structural-recognition failures from assumptions tied to a single script or writing direction.

\begin{table}[ht]
\centering
\caption{Template language, script, and writing direction.}
\label{tab:lang_stats}
\resizebox{\linewidth}{!}{
\begin{tabular}{llccc}
\toprule
Language & Script & Direction & Templates & \% \\
\midrule
Arabic & Arabic & RTL & 8 & 11.43 \\
Chinese & CJK & LTR & 11 & 15.71 \\
Chinese--English & CJK + Latin & LTR & 2 & 2.86 \\
English & Latin & LTR & 19 & 27.14 \\
German & Latin & LTR & 2 & 2.86 \\
Japanese & CJK & LTR & 22 & 31.43 \\
Portuguese & Latin & LTR & 3 & 4.29 \\
Spanish & Latin & LTR & 3 & 4.29 \\
\midrule
\textbf{Total} & & & \textbf{70} & \textbf{100.00} \\
\bottomrule
\end{tabular}
}
\end{table}

\subsection{Instance-level Statistics}
\label{sec:constraint_coverage}

\begin{figure}[t]
\centering
\includegraphics[width=\linewidth]{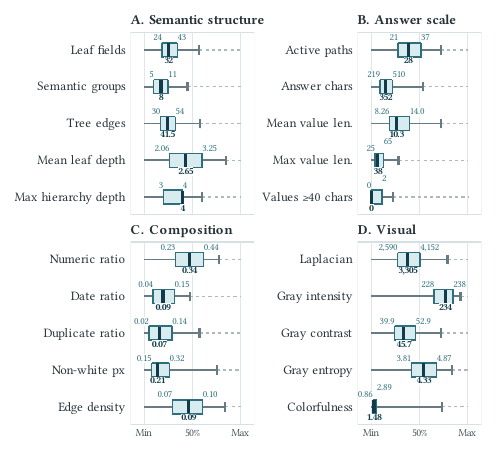}
\caption{Distribution of instance-level semantic structure, answer scale, content composition, and visual properties.}
\label{fig:instance_statistics}
\vspace{-2em}
\end{figure}

\textbf{Q2} examines variation among the 7,000 generated instances. \Cref{fig:instance_statistics} summarizes 20 attributes spanning semantic structure, answer scale, content, and appearance: the median instance contains 32 leaf fields, 8 semantic groups, 41.5 tree edges, 28 active answer paths, and 352 answer characters. Numeric values occupy a median 34\% of populated fields, while the median non-white-pixel ratio and edge density are 21\% and 0.09. The broad distribution tails show that \method varies structural load, response length, value composition, and rendering conditions rather than producing near-duplicate instances.

\Cref{tab:within_template_instance_diversity} isolates variation among instances from the same template. For the 58.6\% of templates with variable schemas, median schema disagreement and optional-path distance reach 0.746 and 0.440; across all templates, median value-set and shared-field differences are 0.775 and 0.673. Visual variation is also universal (median distance 0.094), and 97.1\% of templates have fully unique answers, with a dataset-level uniqueness ratio of 0.989. These row-wise metrics confirm that repeated generation changes structure, content, and appearance, not only field values.

\begin{table}[t]
\centering
\caption{Within-template instance diversity measured by pairwise differences.}
\label{tab:within_template_instance_diversity}
\small
\setlength{\tabcolsep}{4pt}
\renewcommand{\arraystretch}{1.08}
\begin{tabularx}{\linewidth}{@{}>{\raggedright\arraybackslash}Xccc@{}}
\toprule
Metric & Coverage & Median & IQR \\
\midrule
Schema disagreement probability
& 58.6\%
& 0.746
& [0.504, 0.849] \\
Optional-path Hamming distance
& 58.6\%
& 0.440
& [0.325, 0.471] \\
\addlinespace[2pt]
Value-set Jaccard distance
& 100.0\%
& 0.775
& [0.726, 0.831] \\
Shared-field exact-value difference
& 100.0\%
& 0.673
& [0.599, 0.727] \\
\addlinespace[2pt]
Robust visual-feature distance
& 100.0\%
& 0.094
& [0.076, 0.119] \\
\addlinespace[2pt]
Full-answer uniqueness ratio
& 97.1\%
& 1.000
& [1.000, 1.000] \\
\bottomrule
\end{tabularx}
\end{table}

% \noindent{\textbf{Constraint combination coverage.}}
% Diagnostic evaluation of constraint interactions requires that combinations appear in sufficient numbers. Figure~\ref{fig:constraint_heatmap} shows the co-occurrence rates of the six most frequent structural tags against the six visual constraint types. Combinations involving \texttt{local\_grid} and border degradation are the most common (\textcolor{red}{XX}\% of instances), as local-grid templates tend to have denser layouts that naturally produce weak-border conditions under degradation. The least-covered combination is \texttt{reading\_order} with severe geometric distortion (\textcolor{red}{X}\% of instances); this reflects a deliberate choice to limit extreme multi-constraint pressure to a small fraction of the corpus. All single-constraint slices contain at least \textcolor{red}{XXX} test instances, which we treat as the minimum threshold for reliable per-constraint evaluation.

% \begin{figure}[t]
% \centering
% \includegraphics[width=\linewidth]{example-image-duck}
% \caption{Co-occurrence heatmap of structural and visual constraint tags across generated instances. Cell values are percentages of the total instance count. \textcolor{red}{[Replace with actual heatmap.]}}
% \label{fig:constraint_heatmap}
% \end{figure}

Table~\ref{tab:difficulty_dist} shows that L2 and L3 each account for 34.29\% of the corpus and 36.36\% of the test split, while L1 and L4 retain the low- and high-difficulty extremes. Mean form-structure and context-complexity scores rise from 0.3747/0.2511 at L1 to 0.7452/0.3720 at L4, supporting the intended tier progression. Because the Director derives these tiers from assigned constraints, each sample carries its difficulty label without additional annotation.

\begin{table}[t]
\centering
\caption{Difficulty tier distribution across the full corpus and test split.}
\label{tab:difficulty_dist}
\resizebox{\linewidth}{!}{
\begin{tabular}{llccrrrr}
\toprule
Tier & Name & $D_{\mathrm{main}}$ Range & Corpus (\%) & Test (\%) & Mean $S_{\mathrm{form}}$ & Mean
  $C_{\mathrm{context}}$ \\
  \midrule
  L1 & Easy   & $[0, 0.6927)$           & 15.71 &
  9.09  & 0.3747 & 0.2511 \\
  L2 & Medium & $[0.6927, 0.8220)$      & 34.29 &
  36.36 & 0.4460 & 0.3000 \\
  L3 & Hard   & $[0.8220, 1.0273)$      & 34.29 &
  36.36 & 0.5601 & 0.3590 \\
  L4 & Expert & $[1.0273, 2.0000]$      & 15.71 &
  18.18 & 0.7452 & 0.3720 \\
     \bottomrule
\end{tabular}
}
\end{table}

\subsection{Test Split Design and Distributional Coverage}
\label{sec:test_split}

\begin{figure}[ht]
\centering
\includegraphics[width=0.9\linewidth]{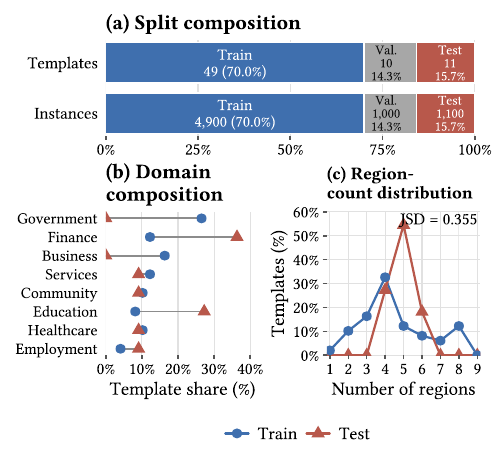}
\caption{Template-disjoint split composition and distributional coverage. }
\label{fig:split_dist}
\end{figure}

\textbf{Q3} asks how a template-disjoint partition balances leakage prevention with evaluation coverage. As shown in Figure~\ref{fig:split_dist}(a), we assign entire templates, rather than individual instances, to the training, validation, and test splits. The resulting 49/10/11-template partition corresponds to 4,900/1,000/1,100 instances, or 70.0\%/14.3\%/15.7\% of the corpus, because every template contributes 100 instances. This construction prevents the same layout from appearing in both training and evaluation. All 1,100 test instances pass automated verification and additionally receive human review.

The template-disjoint split favors leakage control over proportional domain matching. Figure~\ref{fig:split_dist}(b) shows that government and immigration and business and operations account for 26.53\% and 16.33\% of the training templates, respectively, but have no test templates. Conversely, finance, banking, and utilities increases from 12.24\% in training to 36.36\% in test, while education and student services increases from 8.16\% to 27.27\%. The maximum train--test gap is therefore 26.53 percentage points. Since the test split contains only 11 templates, each template changes a domain share by 9.09 percentage points; aggregate test results should thus be interpreted as performance on the represented domains rather than as corpus-wide domain estimates.

Figure~\ref{fig:split_dist}(c) reveals a complementary concentration in structural scale. The training split spans one to eight regions per template, whereas the test split contains only four to six; six of its eleven templates (54.55\%) contain five regions. The resulting Jensen--Shannon divergence of 0.355 indicates a clear separation between the two discrete distributions. Consequently, the test split provides a fully reviewed, template-leakage-free diagnostic subset for mid-range region complexity, but it does not measure generalization to the low- and high-region-count tails. We therefore bound domain- and complexity-level conclusions to the coverage visible in Figure~\ref{fig:split_dist} rather than claiming distributional representativeness.

%% file: sections/evaluation.tex
\section{Evaluation Protocol}
\label{sec:evaluation}

\method evaluates predictions at three complementary levels: page-level metrics measure schema and content recovery, path-level metrics test hierarchical field binding, and component metrics assess regions and line-item groups. Three structure-specific diagnostics additionally evaluate region-local grids, widget groups, and typed relations. We report the levels separately to distinguish recognition, binding, localization, and organization errors.

\subsection{Page-level Metrics}
\label{sec:metrics}

Let $\hat{A}$ and $A^*$ denote the predicted and ground-truth answer JSON objects for a page.

\noindent{\textbf{Schema Normalized Tree Edit Similarity (Schema-nTED).}}
Schema-nTED compares schema trees $T(A)$ that retain field names, container types, and hierarchy while discarding values:
\begin{equation}
  \mathrm{Schema\text{-}nTED}
  = 1 - \frac{d_{\mathrm{APTED}}\!\left(T(\hat{A}),T(A^*)\right)}
  {|T(\hat{A})|+|T(A^*)|}.
\label{eq:schema_nted}
\end{equation}
It equals 1 only when the predicted and ground-truth schemas match, independently of recognized values.

\noindent{\textbf{Value Normalized Edit Similarity (Value-nED).}}
Let $\hat{V}$ and $V^*$ be the multisets of non-empty leaf values in $\hat{A}$ and $A^*$. Value-nED uses normalized Levenshtein similarity $s$ and maximum-weight one-to-one matching:
\begin{equation}
  \mathrm{Value\text{-}nED}
  = \frac{\max_{M\in\mathcal{B}(\hat{V},V^*)}
  \sum_{(\hat{v},v^*)\in M}s(\hat{v},v^*)}
  {\max(|\hat{V}|,|V^*|)},
\label{eq:value_ned}
\end{equation}
where $\mathcal{B}(\hat{V},V^*)$ denotes the set of bipartite matchings. Unmatched values score zero, penalizing missing and spurious content independently of field paths.

\subsection{Component-level Metrics}
\label{sec:schema_metric}

\noindent{\textbf{Table Structure Recognition Path Accuracy (TSR-path).}}
Let $P(A)$ be the set of leaf field-path--value pairs obtained by flattening an answer tree. TSR-path counts a ground-truth leaf only when its complete path and value are reproduced exactly:
\begin{equation}
  \mathrm{TSR\text{-}path}
  = \frac{\sum_{(p,v)\in P(A^*)}
  \mathbbm{1}\!\left[(p,v)\in P(\hat{A})\right]}
  {|P(A^*)|}.
\label{eq:tsr_path}
\end{equation}
It is therefore strict ground-truth leaf recall rather than a precision-aware F1 score.

\noindent{\textbf{Region Detection F1 at IoU 0.5 (R-F1@0.5).}}
Let $\hat{R}$ and $R^*$ denote the predicted and ground-truth region sets, and let $\mathcal{M}_R$ be the maximum one-to-one matching over pairs with the same semantic type and box IoU of at least 0.5. We compute
\begin{equation}
  \mathrm{R\text{-}F1@0.5}
  = \frac{2|\mathcal{M}_R|}{|\hat{R}|+|R^*|}.
\label{eq:rfone}
\end{equation}
Thus, a region requires both correct type and location.

\noindent{\textbf{Line-item-group F1 (LIG-F1).}}
Let $\hat{G}$ and $G^*$ denote predicted and ground-truth line-item-group boxes, and let $\mathcal{M}_G$ be their maximum one-to-one matching at an IoU threshold of 0.5. LIG-F1 is
\begin{equation}
\mathrm{LIG\text{-}F1}
  = \frac{2|\mathcal{M}_G|}{|\hat{G}|+|G^*|}.
\label{eq:lig_f1}
\end{equation}
It measures whether repeated line-item blocks are localized as coherent groups.

\noindent{\textbf{Local-grid GriTS Topology ($\mathrm{LG\text{-}GriTS}_{\mathrm{Top}}$).}}
Let $\hat{\mathcal{G}}$ and $\mathcal{G}^*$ be the predicted and ground-truth region-local grids. We match grids one-to-one only when their parent regions are paired by $\mathcal{M}_R$, using $s_{\mathrm{Top}}=\mathrm{GriTS}_{\mathrm{Top}}$~\cite{grits2023} as the pair weight. For the maximum-weight matching $\mathcal{M}_{\mathrm{LG}}$,
\begin{equation}
  \mathrm{LG\text{-}GriTS}_{\mathrm{Top}}
  = \frac{2\!\sum_{(\hat{g},g^*)\in\mathcal{M}_{\mathrm{LG}}}
  s_{\mathrm{Top}}(\hat{g},g^*)}
  {|\hat{\mathcal{G}}|+|\mathcal{G}^*|}.
\label{eq:lg_grits_top}
\end{equation}
This score isolates row/column topology and span recovery within local grids.

\noindent{\textbf{Widget Grouping F1 (WG-F1).}}
Let $\hat{\mathcal{W}}$ and $\mathcal{W}^*$ denote widget groups and $\mathcal{M}_W$ their maximum valid one-to-one matching. A valid group match requires the same parent field, widget family, option membership, and observable selection states after member alignment. We compute
\begin{equation}
  \mathrm{WG\text{-}F1}
  = \frac{2|\mathcal{M}_W|}{|\hat{\mathcal{W}}|+|\mathcal{W}^*|}.
\label{eq:wg_f1}
\end{equation}

\noindent{\textbf{Typed Relation F1 (Rel-F1).}}
Let $\hat{\mathcal{E}}$ and $\mathcal{E}^*$ be the predicted and ground-truth directed, typed relation-edge sets. After aligning their endpoint objects, $\mathcal{M}_E$ contains edges whose source, target, direction, and relation type all agree. Rel-F1 is
\begin{equation}
  \mathrm{Rel\text{-}F1}
  = \frac{2|\mathcal{M}_E|}{|\hat{\mathcal{E}}|+|\mathcal{E}^*|}.
\label{eq:rel_f1}
\end{equation}

\subsection{Aggregation and Reporting Protocol}

All metrics are macro-averaged over pages. Missing, unparsable, or schema-invalid predictions receive zero on applicable metrics, whereas pages without the corresponding ground-truth component are excluded only from that component-level average. We report valid-output rate separately and use valid-only scores solely as coverage diagnostics. Detailed tree construction, normalization, matching, and edge-case rules are provided in \Cref{app:metric_details}.

%% file: sections/experiment.tex
\section{Experiments}
\label{sec:experiments}

We evaluate 14 API-hosted and locally deployable systems plus two SFT variants to
characterize the current state of table-form document structure
recognition. We first report overall performance under the
hierarchical evaluation protocol, and then analyze how model
behavior varies across difficulty levels, structural constraints,
visual degradations, and evaluation criteria. These analyses are
designed to expose which structural capabilities remain unresolved
and where document-level recovery ceases to be a reliable indicator of fine-grained structural correctness. Our code is publicly available at \githublogo\,\url{https://github.com/D2I-CUHKSZ/FormStruct-Bench}.

\subsection{Baselines}
\label{sec:exp_baselines}

We compare six API-hosted VLMs with eight locally deployable systems spanning open-weight VLMs, document/OCR parsers, and TSR pipelines, and additionally report SFT variants of Qwen3.5-9B and Qwen3.6-35B-A3B. This grouping contrasts vendor-hosted capability with reproducible local alternatives; model roles and schema mappings are detailed in \Cref{app:baseline_details}.

\subsection{Implementation Protocol}
\label{sec:exp_protocol}

All systems are evaluated on the same page images and normalized to the \method JSON schema before scoring. VLMs share the same task definition and use deterministic decoding when available, while local pipelines follow their released inference procedures; parsing, repair, and conversion rules are fixed across systems (\Cref{app:implementation_details}). The prompt is provided in \Cref{app:prompt-text}.

\subsection{Overall Performance and Cross-level Gap}
\label{sec:exp_main_results}

\noindent \textbf{Finding} \circledchar[black]{1}\textbf{: Under our hierarchical evaluation framework, current systems recover document content substantially better than the structures that organize it.}
As summarized in \Cref{tab:main_results}, the best document-level scores reach 83.85 Value-nED and 57.45 Schema-nTED, whereas the maxima on the primary structure metrics are only 17.91 TSR-path, 12.49 R-F1@0.5, and 14.21 LIG-F1. The structure-specific diagnostics remain lower still, peaking at 5.56 LG-GriTS$_{\mathrm{Top}}$, 3.94 WG-F1, and 2.81 Rel-F1. This cross-level gap answers our first question: existing systems can read form content, but they still struggle to bind it to the correct hierarchy, region, local grid, widget group, and relation endpoint.

\noindent \textbf{Finding} \circledchar[black]{2}\textbf{: SFT improves content recovery, region localization, and grouping, but does not uniformly improve all structural capabilities.}
Qwen3.6-35B-A3B-SFT achieves the best Schema-nTED (57.45), R-F1@0.5 (12.49), LIG-F1 (14.21), and WG-F1 (3.94), while Qwen3.5-9B-SFT leads Value-nED (83.85). However, relative to its base model, Qwen3.6-35B-A3B-SFT drops from 17.91 to 16.84 on TSR-path and from 5.56 to 1.00 on LG-GriTS$_{\mathrm{Top}}$; its 2.24 Rel-F1 also remains below Seed 2.1 Pro's 2.81. The uneven gains indicate that supervised adaptation strengthens selected output components rather than resolving the broader coupling problem among document reading, spatial grounding, topology reconstruction, and relation binding.

\newcommand{\modelicon}[1]{\raisebox{-0.35ex}{\includegraphics[height=1.8ex]{#1}}\hspace{0.3em}}

\begin{table*}[ht]
\centering
\caption{Main results on the five primary metrics and three structure-specific diagnostics of \method. All values are percentages. Bold marks the best reported result in each column.}
\label{tab:main_results}
\resizebox{\textwidth}{!}{
\begin{tabular}{@{}llcccccccc@{}}
\toprule
\multirow{2}{*}{Model} & \multirow{2}{*}{Model Size} & \multicolumn{2}{c}{Document-level Metrics} & \multicolumn{3}{c}{Primary Structure Metrics} & \multicolumn{3}{c}{Structure-specific Diagnostics} \\
\cmidrule(lr){3-4}\cmidrule(lr){5-7}\cmidrule(lr){8-10}
& & Schema-nTED $\uparrow$ & Value-nED $\uparrow$ & TSR-path $\uparrow$ & R-F1@0.5 $\uparrow$ & LIG-F1 $\uparrow$ & LG-GriTS$_{\mathrm{Top}}$ $\uparrow$ & WG-F1 $\uparrow$ & Rel-F1 $\uparrow$ \\
\midrule
\multicolumn{10}{@{}l}{\textit{API-hosted}} \\
\modelicon{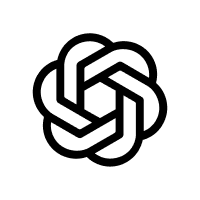}GPT-5.5~\cite{openai_models_2026} & Undisc. & 54.15 & 80.98 & 17.15 & 1.74 & 2.07 & 0.53 & 0.02 & 0.09 \\
\modelicon{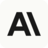}Claude Sonnet 5~\cite{claude_sonnet5_2026} & Undisc. & 52.22 & 75.42 & 13.77 & 1.75 & 0.52 & 1.77 & 0.02 & 0.13 \\
\modelicon{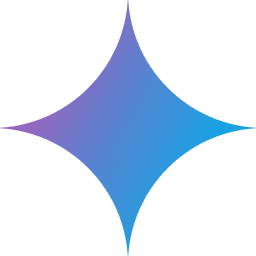}Gemini 3.5 Flash~\cite{gemini_models_2026} & Undisc. & 48.59 & 66.80 & 14.55 & 8.19 & 4.21 & 4.26 & 1.99 & 1.87 \\
\modelicon{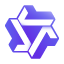}Qwen3.7-Plus~\cite{qwen_models_2026} & Undisc. & 34.01 & 74.02 & 9.75 & 0.00 & 0.00 & 0.00 & 0.00 & 0.00 \\
\modelicon{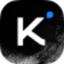}Kimi2.6~\cite{kimi_platform_2026} & Undisc. & 48.20 & 76.63 & 12.96 & 0.47 & 0.19 & 0.00 & 0.05 & 0.09 \\
\modelicon{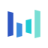}Seed 2.1 Pro~\cite{seed21_model_card_2026} & Undisc. & 52.81 & 77.45 & 15.90 & 10.77 & 12.10 & 5.43 & 0.86 & \textbf{2.81} \\

\midrule
\multicolumn{10}{@{}l}{\textit{Locally deployable}} \\
\modelicon{figures/logos/qwen_logo.png}Qwen3.6-35B-A3B~\cite{qwen36_35b_a3b} & 35B / 3B act. & 48.46 & 66.93 & \textbf{17.91} & 5.95 & 7.65 & \textbf{5.56} & 0.40 & 0.00 \\
\modelicon{figures/logos/qwen_logo.png}Qwen3.6-35B-A3B-SFT~\cite{qwen36_35b_a3b} & 35B / 3B act. & \textbf{57.45} & 77.83 & 16.84 & \textbf{12.49} & \textbf{14.21} & 1.00 & \textbf{3.94} & 2.24 \\
\modelicon{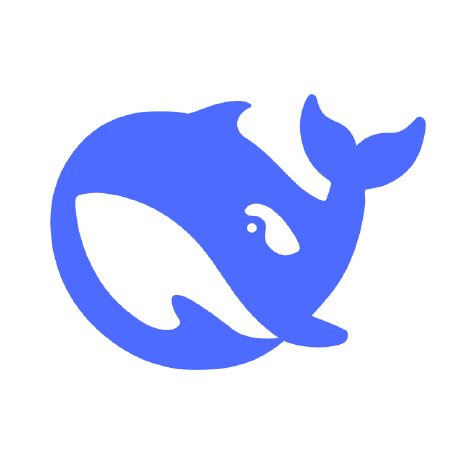}DeepSeek-VL-2~\cite{deepseekocr2_2026} & 27B & 5.89 & 5.29 & 0.00 & 0.00 & 0.01 & 0.00 & 0.00 & 0.00 \\
\modelicon{figures/logos/qwen_logo.png}Qwen3.5-9B~\cite{qwen35_9b} & 9B & 45.58 & 65.73 & 12.58 & 2.03 & 1.84 & 1.23 & 0.01 & 0.00 \\
\modelicon{figures/logos/qwen_logo.png}Qwen3.5-9B-SFT~\cite{qwen35_9b} & 9B & 56.98 & \textbf{83.85} & 17.78 & 10.81 & 9.00 & 0.00 & 0.00 & 0.00 \\
\modelicon{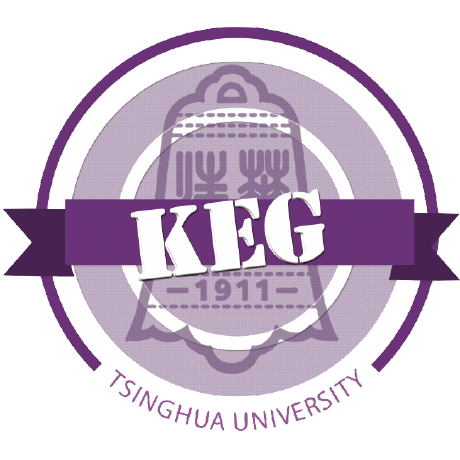}GLM-4.6V-Flash~\cite{glm46v_model_card_2025} & 9B & 30.41 & 47.88 & 4.92 & 5.59 & 3.66 & 1.16 & 0.05 & 0.00 \\
\modelicon{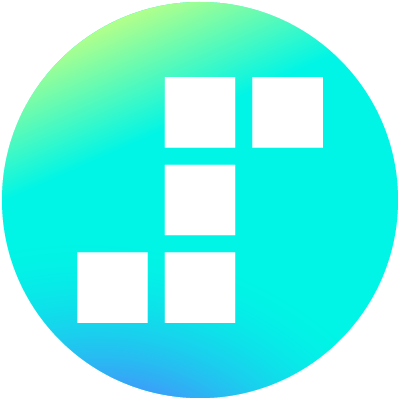}Step3-VL-10B~\cite{huang2026step3vl10b} & 10B & 26.20 & 59.48 & 0.04 & 0.57 & 1.05 & 0.00 & 0.03 & 0.00 \\
\modelicon{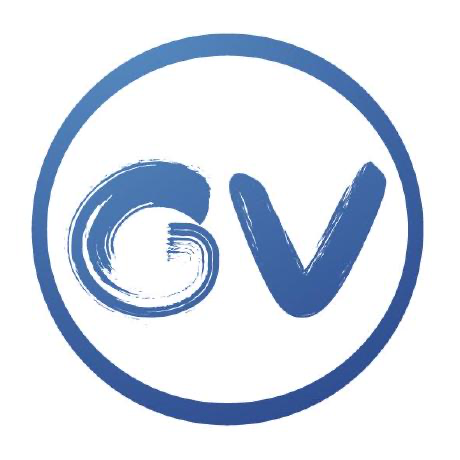}InternVL3.5-8B~\cite{wang2025internvl35} & 8B & 30.34 & 57.03 & 0.01 & 1.05 & 0.00 & 0.00 & 0.01 & 0.00 \\
\modelicon{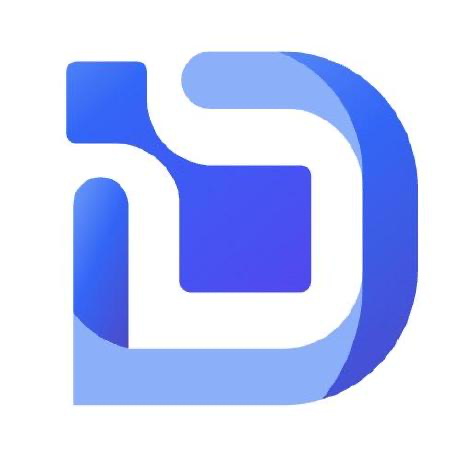}MinerU 2.5 Pro~\cite{mineru2024} & 1.2B & 4.53 & 10.97 & 0.00 & 0.00 & 0.00 & 0.00 & 0.00 & 0.00 \\
\modelicon{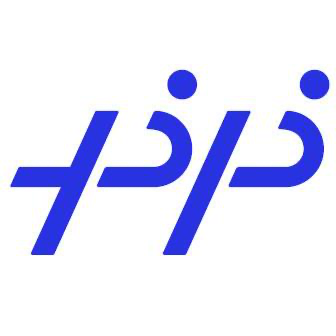}PaddleOCR-VL-1.6~\cite{zhang2026paddleocrvl16expandingfrontierdocument} & 0.9B & 2.97 & 21.73 & 0.00 & 3.68 & 0.00 & 0.00 & 0.00 & 0.00 \\
\bottomrule
\end{tabular}
}
\end{table*}

\subsection{Difficulty Analysis}
\label{sec:exp_difficulty}
\begin{figure}[ht]
    \centering
    \includegraphics[width=0.85\linewidth]{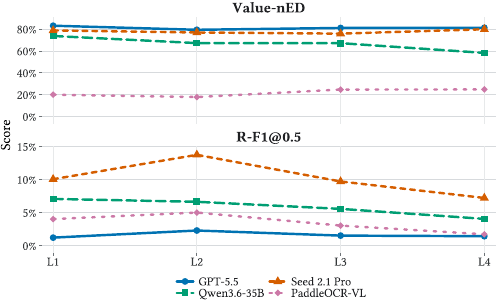}
    \caption{Performance across Difficulty L1--L4.}
    \label{fig:difficulty_boundary}
    \vspace{-2em}
\end{figure}

\noindent \textbf{Finding} \circledchar[black]{3}\textbf{: Increasing structural difficulty widens the gap
between content recovery and spatial grounding.} As shown in \Cref{fig:difficulty_boundary}, across the L1--L4 tiers, document-level value recovery is generally
more stable than region-level structural localization. GPT-5.5 and
Seed 2.1 Pro maintain comparatively stable Value-nED, whereas
Seed 2.1 Pro's R-F1@0.5 decreases from 13.76 at L2 to 7.23 at
L4, a 47.5\% relative reduction. Qwen3.6-35B-A3B exhibits
degradation at both levels, but its region score drops more sharply:
Value-nED decreases from 73.85 at L1 to 58.37 at L4, while
R-F1@0.5 falls from 7.09 to 4.07, corresponding to relative
reductions of 21.0\% and 42.6\%, respectively. This asymmetric
degradation suggests that current multimodal models can continue
recovering plausible field content from global semantic cues as
structural and contextual complexity increases, but struggle to
preserve precise region identities and bind that content to the
correct local organization. The near-flat R-F1@0.5 trajectory of
GPT-5.5 should not be interpreted as robustness, since the model
already operates close to the metric floor.

\subsection{Visual Degradation Robustness}
\label{sec:visual_analysis}
\noindent \textbf{Finding} \circledchar[black]{4}\textbf{:  Visual degradation leaves the page-level capability hierarchy largely intact but consistently weakens exact hierarchical binding, while region localization and grouping exhibit model-specific responses.}
\begin{figure}[ht]
    \centering
    \includegraphics[width=0.9\linewidth]{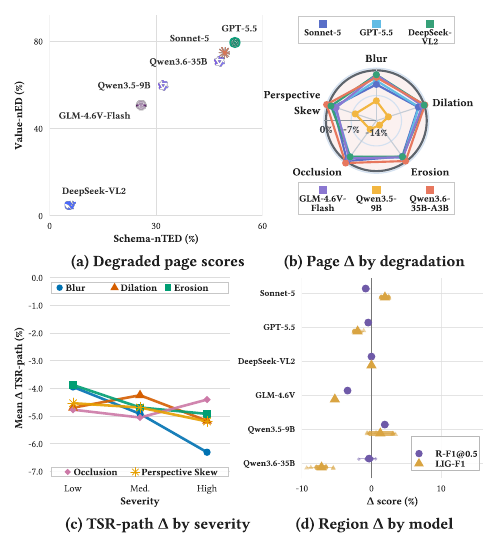}
    \caption{Visual-degradation robustness: (a) degraded page-level scores; (b) page-level changes by degradation type; (c)TSR-path changes by severity; and (d) region-level changes by model.}
    \label{fig:visual_degradation}
    \vspace{-1.5em}
\end{figure}
Visual degradation exposes a clear capability boundary in current
multimodal models. Their page-level extraction can remain relatively stable because visible values and document schemas can be recovered from redundant textual and global semantic cues. In contrast, exact hierarchical binding depends more heavily on fragile local evidence, such as boundaries, alignment, spacing, and the visual correspondence between labels and fields. Region localization and line-item grouping are also not affected uniformly, indicating that detecting an approximate document region and preserving the internal grouping of its elements are only partially coupled capabilities. These results suggest that current models lack a visual-structural representation that remains consistently aligned across semantic extraction, hierarchical binding, localization, and grouping when document appearance changes.

\Cref{fig:visual_degradation} supports this boundary. The page-level ranking remains largely unchanged under degradation (\Cref{fig:visual_degradation}(a)), and stronger models show only limited degraded--clean changes (\Cref{fig:visual_degradation}(b)). In contrast, TSR-path decreases under every degradation type and severity level, with losses of approximately 3.8--6.3 points
(\Cref{fig:visual_degradation}(c)), showing that exact value-to-path binding is consistently more fragile than page-level extraction. Region-level effects are less uniform (\Cref{fig:visual_degradation}(d)): some models preserve region
localization while losing line-item grouping, indicating that localization and grouping remain only partially coupled. Overall, visual degradation primarily destabilizes fine-grained structural binding rather than coarse document semantics.
\subsection{In-Domain Adaptation \& Cross-Dataset Transfer}
\label{sec:exp_diagnostic_validation}
\begin{figure}[ht]
    \centering
    \includegraphics[width=0.9\linewidth]{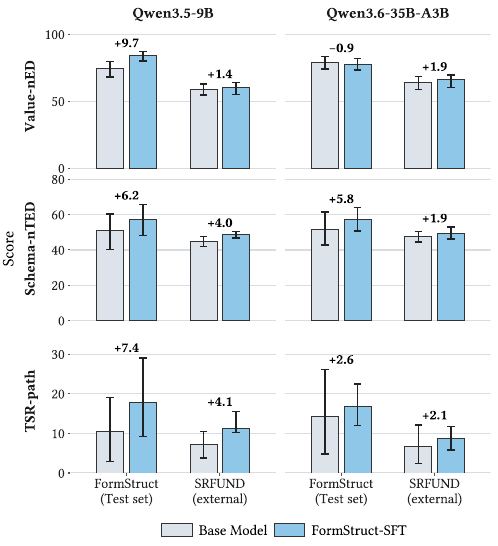}
    \caption{Models fine-tuned on synthetic table-form documents achieve consistent gains on real-world data.}
    \label{fig:sft_transfer}
    \vspace{-2em}
\end{figure}

\noindent \textbf{Finding} \circledchar[black]{5}\textbf{: SFT on synthetic table-form documents improves real-world structure recognition across both model scales, covering content recovery, schema reconstruction, and hierarchical value binding.}
To test whether the generated data provides transferable supervision rather than merely fitting the synthetic distribution, we fine-tune two model scales on synthetic table-form documents and evaluate them on both synthetic and real-world data. As shown in \Cref{fig:sft_transfer}, SFT improves all three metrics on real documents for both models, covering content recovery, schema reconstruction, and hierarchical value binding. The gains in Schema-nTED and TSR-path further show that the models learn structural organization and field-path alignment, not only surface-text recognition. These results indicate that the synthetic data captures transferable structural regularities and can effectively support real-world table-form structure recognition.

% Relative to relaxed scoring, full criteria reduce representative API and local scores by 58.1--63.4\% and 42.0--51.8\%, respectively (Figure~\ref{fig:evaluation_sensitivity}). Relation penalties are stable across groups, whereas local-grid effects vary by model.

% \begin{table}[ht]
% \centering
% \caption{TSR-path loss (\%) by constraint for four local VLMs.}
% \label{tab:constraint_errors}
% \small
% \begin{tabular*}{0.90\linewidth}{@{\extracolsep{\fill}}lccc@{}}
% \toprule
% Constraint & Min & Mean & Max \\
% \midrule
% Dense key--field relations & 7.2 & 11.5 & 15.7 \\
% Region-local grids & 3.5 & 10.8 & 17.6 \\
% Widget grouping & 4.5 & 6.3 & 8.3 \\
% Mixed layout & $-1.2$ & 3.3 & 13.0 \\
% Line-item groups & $-5.0$ & $-0.9$ & 1.8 \\
% \bottomrule
% \end{tabular*}
% \end{table}

% Table~\ref{tab:constraint_errors} identifies dense relations and local grids as the largest mean losses (11.5/10.8 points), followed by widgets (6.3). Mixed-layout and line-item effects are inconsistent, localizing errors to field binding and local organization.

%% file: sections/related_work.tex
\section{Related Work}

\noindent\textbf{Table structure recognition.}
Existing TSR benchmarks mainly study regular or scientific tables under a global-grid view.
TableBank and SciTSR cover table detection and scientific-table structure recognition \cite{tablebank2020,scitsr2019}, while PubTabNet introduces TEDS to measure tree-edit similarity over HTML-like tables \cite{pubtabnet2020}.
PubTables-1M scales scientific-document table extraction with canonicalized annotations \cite{pubtables1m2021}, and GriTS evaluates topology, content, and location similarity \cite{grits2023}.
These benchmarks omit form-specific elements such as fillable fields, option groups, local subgrids, and non-local relations; consequently, table-level scores can obscure failures in localization, span recovery, widget grouping, reading order, and local-grid reconstruction.

\noindent\textbf{Document AI and form-like documents.}
Full-page layout and parsing benchmarks, including PubLayNet, DocBank, DocLayNet, and OmniDocBench \cite{publaynet2019,docbank2020,doclaynet2022,omnidocbench2025}, are supported by general document-understanding systems such as the LayoutLM family, LayoutXLM, Donut, and MinerU \cite{layoutlm2020,layoutlmv22021,layoutlmv32022,layoutxlm2021,donut2022,mineru2024}.
Form and receipt datasets such as FUNSD, SROIE, and DocILE instead focus on key--value and business-information extraction \cite{funsd2019,sroie2021,docile2023}.
SRFUND directly benchmarks multi-granularity hierarchical form reconstruction \cite{ma2024srfund}, while \method emphasizes explicit table-form structures and diagnostic evaluation.
VAREX closely overlaps with our template-based synthesis, deterministic ground truth, and quality-assurance workflow \cite{barzelay2026varex}, but targets schema-based value extraction rather than structure reconstruction.
These tasks may recover correct values without reconstructing the local grids, widget groups, section membership, and relations that make forms machine-consumable.
\method therefore complements them with an explicit hierarchy over regions, local grids, fields, widgets, and relations, evaluated through component-level diagnostics rather than a single table or extraction score.

%% file: sections/conclusion.tex
\section{Conclusion}

\method reframes table-form document structure recognition as hierarchical recovery over semantic regions, local grids, fields, widgets, and typed relations.
Its 70 templates and 7,000 verified instances support five complementary metrics and diagnostic slices over difficulty, structural constraints, and visual degradation.
Across 14 systems and two SFT variants, document-level performance reaches 83.85\%, yet the strongest fine-grained structural score remains below 18\%, with no system dominating all components.
Within our template-disjoint test coverage, this gap shows that recovering content does not imply recovering the hierarchy required for reliable table-form understanding.
\method thus measures progress at the level of structural failure rather than aggregate output similarity.

%% file: sections/appendix.tex
\section{Additional Experimental Details}
\label{app:experimental_details}
\providecommand{\tbd}[1]{\textcolor{red}{\textbf{TBD:} #1}}

\subsection{Baseline Systems}
\label{app:baseline_details}

The API-hosted group comprises GPT-5.5~\cite{openai_models_2026}, Claude Sonnet 5~\cite{claude_sonnet5_2026}, Gemini 3.5 Flash~\cite{gemini_models_2026}, Qwen3.7-Plus~\cite{qwen_models_2026}, Kimi2.6~\cite{kimi_platform_2026}, and Seed 2.1 Pro~\cite{seed21_model_card_2026}. These systems directly generate the complete target JSON and represent vendor-hosted multimodal models available only through remote inference services.

The locally deployable VLMs include Qwen3.6-35B-A3B~\cite{qwen36_35b_a3b}, DeepSeek-VL-2~\cite{deepseekocr2_2026}, Qwen3.5-9B~\cite{qwen35_9b}, and GLM-4.6V-Flash~\cite{glm46v_model_card_2025}. We additionally evaluate Step3-VL-10B~\cite{huang2026step3vl10b} and InternVL3.5-8B~\cite{wang2025internvl35}. The locally deployable document/OCR pipelines are MinerU 2.5 Pro~\cite{mineru2024} and PaddleOCR-VL-1.6. The VLMs are prompted to generate the target schema, whereas the document/OCR pipelines retain their released output interfaces. For the latter, only supported predictions are mapped into the closest compatible subset of the \method schema; components not produced by a pipeline remain absent after conversion.

\subsection{Experimental Environment}
\label{app:experimental_environment}

\paragraph{Hardware.}
Local inference was conducted on a server with two NVIDIA A100-SXM4 GPUs (80~GB each; 81,920~MiB reported per device) using driver 550.90.07, with Multi-Instance GPU (MIG) disabled. The preserved host has two 48-core AMD EPYC 7A23 CPUs (96 physical cores and 192 logical CPUs) and approximately 1~TiB of RAM, and runs Debian GNU/Linux~13 user space on a Linux~5.15 kernel. The surviving inference logs corroborate the GPU model and driver for the formal runs; CPU, memory, and operating-system details were not stored in per-run manifests and are reported from the preserved host snapshot.

\paragraph{Software.}
Table~\ref{tab:software_environment} summarizes the package versions that can be verified from preserved environments and formal server logs. The NVIDIA driver reports CUDA~12.4 compatibility, whereas the main vLLM stack uses CUDA~12.8 wheels and the Paddle stack uses CUDA~12.6 wheels; we therefore report these layers separately rather than assigning a single CUDA version to all runs.

\begin{table}[h]
\centering
\caption{Verified software environments for locally deployable baselines.}
\label{tab:software_environment}
\small
\setlength{\tabcolsep}{4pt}
\begin{tabularx}{\columnwidth}{@{}l>{\raggedright\arraybackslash}X@{}}
\toprule
Stack & Verified environment \\
\midrule
vLLM & Python 3.12.8; PyTorch 2.10.0+cu128; Transformers 4.57.6; vLLM 0.18.0. \\
MinerU & MinerU-VL-Utils 1.0.5 with its in-process vLLM inference engine. \\
PaddleOCR-VL & SGLang 0.5.2; PyTorch 2.8.0; Transformers 4.56.1; PaddlePaddle-GPU 3.2.1; PaddleOCR 3.7.0; PaddleX 3.7.2. \\
\bottomrule
\end{tabularx}
\end{table}

\subsection{Inference and Output Normalization}
\label{app:implementation_details}

Each VLM receives the original page image and the same instruction to return a JSON object containing a semantic answer tree, regions, widgets, local grids, cells, and relations. The shared prediction prompt is reproduced verbatim in \Cref{app:prompt-text}. We use deterministic or near-deterministic decoding whenever the API or inference backend exposes the corresponding control. Locally deployable document/OCR and TSR systems are run with their released image-to-table or document-parsing procedures.

Predictions are processed by the same task-agnostic JSON parser and schema-conversion pipeline. The parser may repair invalid JSON syntax, but it does not infer or manually complete missing fields; any field absent after parsing remains missing during evaluation. Native pipeline outputs are then converted into their compatible \method components before the document- and component-level metrics are computed.
\subsection{Supervised Fine-Tuning Details}
  We performed parameter-efficient supervised fine-tuning (SFT) on the \method training split using LoRA.  Each SFT example consisted of a rendered form image, a task instruction, and a structured JSON target. Only assistant-side JSON tokens contributed to the language-modeling loss; image and instruction tokens were masked.
 Both models were trained for one epoch, corresponding to 307
  optimizer updates. Validation was performed at step 300. After training, the LoRA adapters were merged into their respective base models in bfloat16 precision using safe weight merging.

  \begin{table}[t]
  \centering
  \caption{Hyperparameters used for \method LoRA-based SFT.}
  \label{tab:sft_hyperparameters}
  \small
  \begin{tabularx}{\linewidth}{lXX}
  \toprule
  Hyperparameter & Qwen3.5-9B & Qwen3.6-35B-A3B \\
  \midrule
  Optimization method        & LoRA SFT & LoRA SFT \\
  Epochs / optimizer steps   & 1 / 307 & 1 / 307 \\
  Peak learning rate         & $5\times10^{-5}$ & $5\times10^{-5}$ \\
  Optimizer                   & Fused AdamW & Fused AdamW \\
  Weight decay                & 0 & 0 \\
  LR schedule                 & Cosine, 3\% warmup & Cosine, 3\% warmup \\
  Per-GPU batch size          & 2 & 1 \\
  Gradient accumulation       & 4 & 8 \\
  Effective global batch size & 16 & 16 \\
  Maximum sequence length     & 12,288 & 14,336 \\
  Maximum image pixels        & 1,048,576 & 1,048,576 \\
  LoRA rank $r$               & 16 & 16 \\
  LoRA scaling $\alpha$       & 32 & 32 \\
  LoRA dropout                & 0.05 & 0.05 \\
  LoRA bias                   & None & None \\
  Random seed                 & 42 & 42 \\
  \bottomrule
  \end{tabularx}
  \end{table}

\subsection{Metric Computation and Edge Cases}
\label{app:metric_details}

\paragraph{Schema-nTED.}
Each answer is converted into a schema tree $T(A)$ that retains field names, container types (object, array, or scalar), and hierarchy while discarding field values. Object keys are stripped and sorted so that serialized object order does not affect the score, whereas array order is preserved. In the APTED distance used by \Cref{eq:schema_nted}, insertion and deletion cost 1; renaming costs 0 for identical node labels and 2 otherwise.

\paragraph{Value-nED.}
Before extracting the multisets of non-empty leaf values, we apply Unicode NFKC, lowercasing, whitespace collapsing, and removal of spaces or commas between digits. Pairwise similarity is
\begin{equation}
  s(\hat{v},v^*) = 1 - \frac{d_{\mathrm{Lev}}(\hat{v},v^*)}
  {\max(|\hat{v}|,|v^*|)}.
\label{eq:value_similarity}
\end{equation}
The maximum-weight bipartite matching in \Cref{eq:value_ned} is one-to-one; unmatched predicted or ground-truth values contribute zero.

\paragraph{TSR-path.}
We recursively flatten each answer tree into leaf field-path--value pairs. Strings are stripped and repeated whitespace is collapsed, but no fuzzy matching, case folding, or Unicode normalization is applied. Extra predicted fields do not change the denominator in \Cref{eq:tsr_path}; the metric is therefore strict ground-truth field recall.

\paragraph{Region and line-item-group matching.}
For R-F1@0.5, a region pair is a valid matching candidate only when its semantic types agree and box IoU is at least 0.5. We then compute the maximum-cardinality one-to-one matching $\mathcal{M}_R$ used in \Cref{eq:rfone}. For LIG-F1, region types are not involved: predicted and ground-truth line-item-group boxes are matched one-to-one whenever their IoU is at least 0.5, producing $\mathcal{M}_G$ in \Cref{eq:lig_f1}.

\paragraph{Region-local grid topology.}
For $\mathrm{LG\text{-}GriTS}_{\mathrm{Top}}$, each local grid is evaluated only inside its associated semantic region. Widgets are removed before grid construction. Following GriTS~\cite{grits2023}, a cell with top-left grid position $(\rho,\theta)$, rowspan $\alpha$, and colspan $\beta$ is repeated at every grid position $(i,j)$ it occupies and represented there by the topology box
\begin{equation}
  b^{\mathrm{Top}}_{i,j}
  = [\theta-j,\,\rho-i,\,\theta-j+\beta,\,\rho-i+\alpha].
\label{eq:local_grid_topology_box}
\end{equation}
The official factored two-dimensional most-similar-substructure alignment compares these boxes with IoU, yielding $s_{\mathrm{Top}}\in[0,1]$. We use these scores as weights in a maximum-weight one-to-one matching between predicted and ground-truth grids whose parent regions are already paired by $\mathcal{M}_R$. The denominator in \Cref{eq:lg_grits_top} therefore penalizes omitted and spurious whole grids, while the GriTS term penalizes incorrect rows, columns, and spans within a matched grid. A malformed grid with non-positive spans, conflicting cell occupancy, or no valid cells receives zero.

\paragraph{Widget-group matching.}
A canonical widget group consists of one parent field and all widgets attached to it by field-to-widget relations; explicit group identifiers, when available, must induce the same partition. Raw prediction identifiers are not compared. Parent fields are aligned by their canonical schema paths within matched parent regions, after which member widgets are matched one-to-one at IoU $\geq0.5$. A member match requires the same widget type and, when a ground-truth state is defined, the same selected/unselected state. A widget group contributes to $\mathcal{M}_W$ only when the parent fields correspond and all members admit a bijection, so a missing, extra, mistyped, or state-inconsistent member makes the group incorrect. Visible label text is excluded to keep WG-F1 focused on grouping rather than OCR.

\paragraph{Typed-relation matching.}
Each relation is canonicalized as a directed triple $(u,\ell,v)$, where $\ell$ is one of the released relation types, including key-to-field, field-to-widget, section-membership, and line-item-membership links. We construct a partial endpoint map $\phi$: regions use $\mathcal{M}_R$; fields use exact canonical schema paths within matched regions; widgets use the member matches from WG-F1; local cells use the alignment induced by $\mathcal{M}_{\mathrm{LG}}$; and line-item groups use $\mathcal{M}_G$. A predicted edge $(\hat{u},\hat{\ell},\hat{v})$ is in $\mathcal{M}_E$ only if both endpoints are mapped and $(\phi(\hat{u}),\hat{\ell},\phi(\hat{v}))\in\mathcal{E}^*$. Edge direction and type must match exactly; edges with an unmatched endpoint remain false positives, and duplicate triples are collapsed before scoring.

\paragraph{Aggregation and invalid outputs.}
Metrics are macro-averaged over pages. Missing, unparsable, or schema-invalid predictions receive zero for every applicable metric. If a ground-truth component exists but is omitted from the prediction, its component score is zero; if the ground-truth component itself is absent, that page is excluded only from the corresponding component-level average. In particular, pages without a ground-truth line-item group are marked not applicable rather than counted as perfect negatives. We report valid-output rate separately, and any score conditioned on valid outputs is treated only as a coverage diagnostic.

\subsection{Qualitative Analysis of Relation Errors}
\label{app:relation_error_cases}

\begin{figure*}[t]
  \centering
  \includegraphics[width=\textwidth]{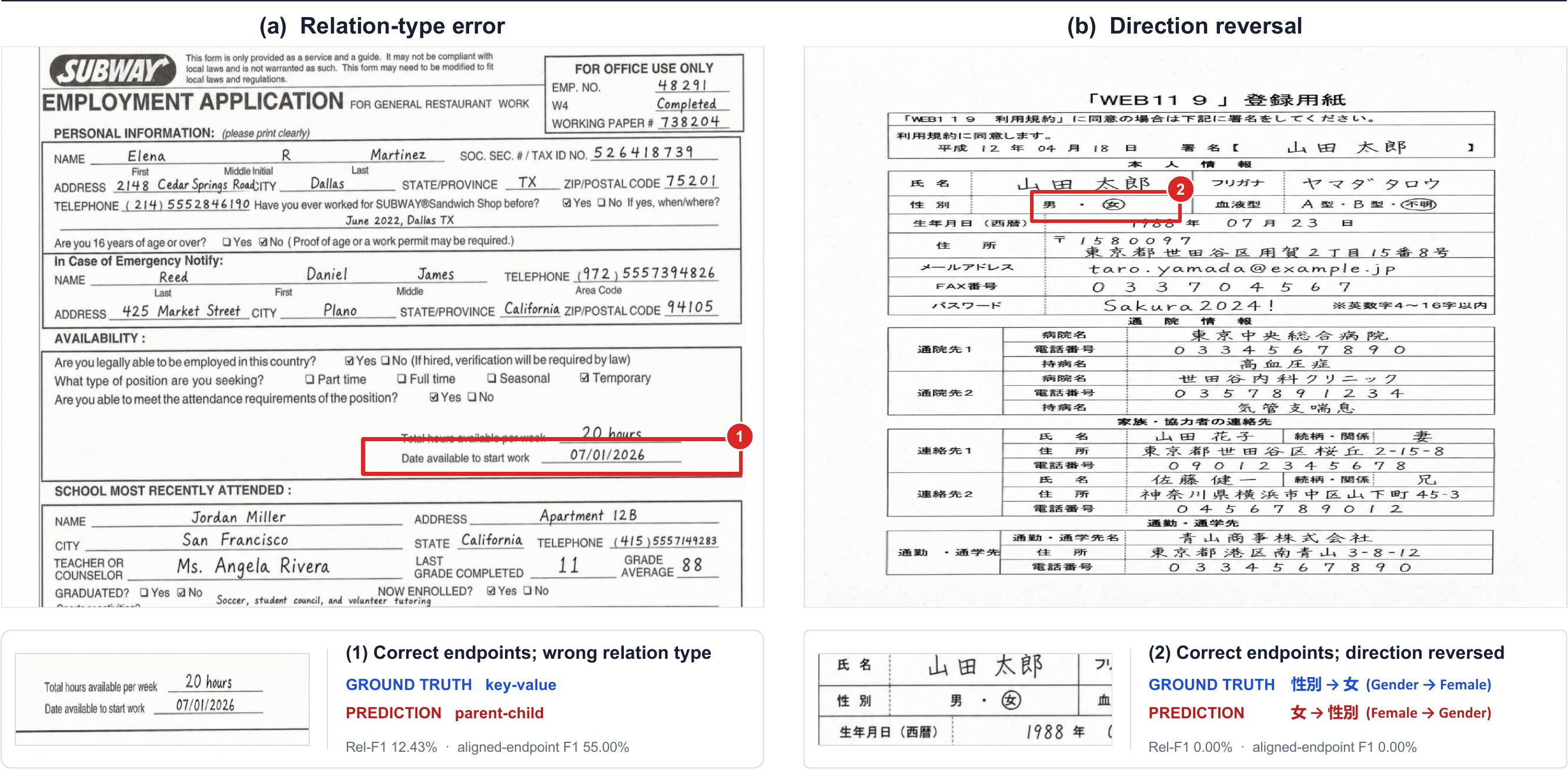}
  \caption{Representative Rel-F1 errors after successful endpoint alignment. Red boxes and bottom insets identify the evaluated relations. Aligned-endpoint F1 is Rel-F1 computed only over edges whose two endpoints can be aligned; its exact type and direction requirements remain unchanged.}
  \Description{Two full-page form examples. The English panel highlights a start-date field and value for which the predicted relation type is parent-child rather than key-value. The Japanese panel highlights a Gender field and Female widget for which the predicted field-widget edge is reversed.}
  \label{fig:relation_error_cases}
\end{figure*}

\Cref{fig:relation_error_cases} separates relation-semantic errors from failures to localize or align their endpoint objects. In the English example, the model links \emph{Date available to start work} to the correct value box and preserves the edge direction, but predicts \texttt{parent-child} instead of the ground-truth \texttt{key-value} type. This type substitution makes the predicted edge a false positive and leaves the corresponding ground-truth edge unmatched. At the page level, 11 of 82 predicted relations match 95 ground-truth relations, yielding 12.43\% Rel-F1; after restricting the comparison to endpoint-alignable relations, the same 11 matches are scored against 19 predictions and 21 ground-truth relations, yielding 55.00\% aligned-endpoint F1.

The Japanese example isolates a complementary failure. The model identifies both the \emph{Gender} field and the selected \emph{Female} widget and predicts the \texttt{field-widget} type, but reverses the directed edge from \emph{Gender $\rightarrow$ Female} to \emph{Female $\rightarrow$ Gender}. The matched, predicted, and ground-truth relation counts are respectively 0, 78, and 75 on the full page, and 0, 4, and 2 within the endpoint-alignable subset, producing 0.00\% under both scores. Together, these examples show that low Rel-F1 is not explained solely by failed endpoint localization: even when the relevant objects are successfully aligned, a model can still confuse the relation ontology or its direction. The cases are illustrative and are not used to estimate error prevalence.

\subsection{External Structural Representativeness and Diagnostic Utility}
\label{app:representativeness}

\paragraph{Real-template grounding and leakage-controlled reference.}
All 7,000 instances in \method are derived from 70 selected, real, manually structured form templates; synthesis changes field content, schema realization, and visual conditions while remaining anchored to a human-created layout skeleton. We evaluate how well these skeletons align with an external real-form reference using SRFUND~\cite{ma2024srfund}. Because pages are not independent layout samples, we cluster SRFUND's 1,592 pages using rotation-aware image hashes, OCR shingles, normalized entity-box occupancy, aspect ratio, and entity-count agreement, obtaining 1,566 layout clusters. Cross-dataset provenance and visual/OCR matching identify 36 confirmed and one corroborated overlap; excluding all 37 affected clusters leaves 1,529 non-overlapping reference clusters. Thus neither repeated pages nor source overlap can inflate the comparison.

\paragraph{Real--Real calibrated support.}
Let $\mathcal{T}$ denote FormStruct templates and $\mathcal{R}$ denote deduplicated SRFUND layout clusters. We compute a group-balanced Gower distance $d_G$ that first averages correlated features within hierarchy, relation, layout, language, entity-composition, and table groups, and then weights the groups equally. Direct-only comparisons retain aligned language, hierarchy, untyped relation, and normalized bounding-box features; the mapped analysis additionally includes explicitly marked conditional mappings for entity composition, table structure, script, and spatial density. In each of 1,000 seeded rounds, two disjoint, size-matched SRFUND sets form the Real--Real baseline, and $\tau$ is the 95th percentile of Real-B-to-Real-A nearest-neighbor distances. We report
\begin{equation}
  \mathrm{TemplateSupport}@\tau =
  \frac{1}{|\mathcal{T}|}\sum_{t\in\mathcal{T}}
  \mathbf{1}\!\left[\min_{r\in\mathcal{R}} d_G(t,r)\leq\tau\right],
  \label{eq:template_support}
\end{equation}
which measures how much of the benchmark template set lies within typical real-layout support rather than whether both corpora have identical sampling frequencies.

\begin{table*}[t]
\centering
\caption{Real--Real calibrated structural support against non-overlapping SRFUND clusters. T--R is the mean template-to-real nearest-neighbor distance; R--R is its size-matched real-to-real baseline. Support is defined in \Cref{eq:template_support}. Higher support and a distance ratio closer to one indicate stronger alignment.}
\label{tab:external_representativeness}
\small
\setlength{\tabcolsep}{5pt}
\begin{tabular}{@{}llrrrrr@{}}
\toprule
Scope & Features / calibration & Matched $n_T/n_R$ & T--R & R--R median [95\% interval] & Ratio & Support (\%) [95\% interval] \\
\midrule
All comparable & Direct only & 70/70 & 0.127 & 0.044 [0.040, 0.050] & 2.89 & 45.7 [31.4, 65.7] \\
Shared languages & Direct only & 60/60 & 0.074 & 0.046 [0.040, 0.052] & 1.61 & 61.7 [45.0, 88.3] \\
Shared languages & Direct, language-stratified & 60/60 & 0.063 & 0.042 [0.038, 0.048] & 1.50 & 78.3 [60.0, 91.7] \\
Shared languages & Direct + conditional mappings & 60/60 & 0.084 & 0.053 [0.047, 0.061] & 1.60 & 70.0 [43.3, 93.3] \\
\bottomrule
\end{tabular}
\end{table*}

The shared-language comparison is the most conservative evidence because it removes unsupported language categories and prioritizes directly aligned features. Under this setting, 61.7\% of templates fall within Real--Real calibrated support, rising to 78.3\% when the calibration preserves language composition. For the complementary reference-to-template direction, we define
\begin{equation}
  \mathrm{ReferenceCoverage}@\tau =
  \frac{1}{|\mathcal{R}|}\sum_{r\in\mathcal{R}}
  \mathbf{1}\!\left[\min_{t\in\mathcal{T}} d_G(r,t)\leq\tau\right].
  \label{eq:reference_coverage}
\end{equation}
Using the global mapped-feature threshold, 876 of 1,529 SRFUND clusters (57.3\%) and 806 of 1,137 shared-language clusters (70.9\%) have a nearby FormStruct template. FormStruct therefore covers a substantial portion of independently collected real layouts without merely reproducing SRFUND's composition. The remaining distance above the Real--Real baseline is expected from a benchmark that deliberately emphasizes denser structures: compared with SRFUND, FormStruct contains more entities and relation links while retaining similar median hierarchy depth, relation density, and normalized entity-box area. We consequently characterize \method as \emph{coverage-oriented rather than prevalence-matched}.

\paragraph{Advantages over the external reference.}
The comparison above evaluates only the intersection of the two annotation schemas. Within that shared space, the real-template grounding and leakage-controlled calibration establish that FormStruct is not an unconstrained synthetic-layout collection. Beyond it, \method explicitly annotates semantic regions, region-local grids and spans, widget types and states, line-item groups, and typed structural relations, none of which has a directly equivalent SRFUND annotation. This combination provides both external grounding and finer diagnostic resolution. Consistent with this purpose, \Cref{tab:main_results} shows that no evaluated system dominates the five primary metrics and that strong value recovery can coexist with near-zero region or grouping scores; the added structural supervision therefore exposes failure modes hidden by document-level extraction alone. This evidence supports the benchmark's diagnostic utility, while claims about synthetic-data augmentation or real-world training gains require a separate transfer experiment. The external comparison is also deliberately scoped to template-level shared structure and does not assert that content, missingness, handwriting, or visual-perturbation frequencies of all generated instances match a real-world population.
\subsection{Evaluation Sensitivity and Error Analysis}
\label{sec:exp_error_analysis}

For each metric $q$ and predefined constraint tag $c$, we summarize the descriptive performance gap as
\begin{equation}
  \Delta_q(c)=\mathbb{E}[q\mid \neg c]-\mathbb{E}[q\mid c].
\label{eq:constraint_gap}
\end{equation}
A positive value indicates lower performance on tagged pages and is interpreted descriptively rather than causally.

\begin{figure}[ht]
    \centering
    \includegraphics[width=0.9\linewidth]{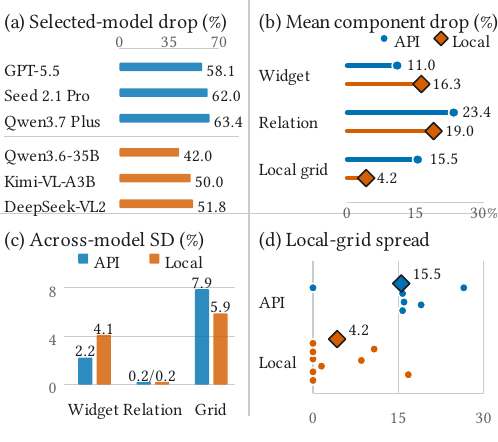}
    \caption{Evaluation sensitivity of API-hosted and local models.}
    \label{fig:evaluation_sensitivity}
\end{figure}

Relative to relaxed scoring, full criteria reduce representative API and local scores by 58.1--63.4\% and 42.0--51.8\%, respectively (Figure~\ref{fig:evaluation_sensitivity}). Relation penalties are stable across groups, whereas local-grid effects vary by model.

\begin{table}[ht]
\centering
\caption{TSR-path loss (\%) by constraint for four local VLMs.}
\label{tab:constraint_errors}
\small
\begin{tabular*}{0.90\linewidth}{@{\extracolsep{\fill}}lccc@{}}
\toprule
Constraint & Min & Mean & Max \\
\midrule
Dense key--field relations & 7.2 & 11.5 & 15.7 \\
Region-local grids & 3.5 & 10.8 & 17.6 \\
Widget grouping & 4.5 & 6.3 & 8.3 \\
Mixed layout & $-1.2$ & 3.3 & 13.0 \\
Line-item groups & $-5.0$ & $-0.9$ & 1.8 \\
\bottomrule
\end{tabular*}
\end{table}

Table~\ref{tab:constraint_errors} identifies dense relations and local grids as the largest mean losses (11.5/10.8 points), followed by widgets (6.3). Mixed-layout and line-item effects are inconsistent, localizing errors to field binding and local organization.

\subsection{Annotation Agreement and Review Outcomes}
\label{app:annotation_quality}

\paragraph{Annotation and verification protocol.}
The 70 templates were divided evenly between two annotators, each of whom annotated 35 templates (50\%). A third annotator reviewed all completed annotations for consistency and correctness. Because the annotation contains both categorical and structured objects, we do not collapse agreement into a single statistic. Categorical decisions are measured with Cohen's $\kappa$, whereas spatial, topological, and relational objects are compared after matching their identities and boundaries. \Cref{tab:annotation_agreement} reports the category-specific statistics obtained during verification. Spatial objects were aligned using maximum-IoU one-to-one bipartite matching, with a pair considered valid when its bounding-box IoU was at least 0.5. Unmatched objects were treated as missing or additional annotations, while region types were compared only after geometric matching.

\begin{table}[t]
\centering
\caption{Agreement between initial template annotations and third-annotator verification.}
\label{tab:annotation_agreement}
\small
\setlength{\tabcolsep}{3pt}
\begin{tabularx}{\columnwidth}{@{}>{\raggedright\arraybackslash}X>{\raggedright\arraybackslash}X>{\raggedright\arraybackslash}X@{}}
\toprule
Target & Agreement statistic & Result \\
\midrule
Region boundaries & Mean IoU; object $F_1$ at IoU $\geq 0.5$ & 0.971,0.903 \\
Region types & Cohen's $\kappa$ on matched regions & 1.000 \\
Local-grid topology & Cell/span topology $F_1$ & 1.000 \\
Widget types and states & Matched-object $F_1$ & 1.000 \\
Structural relations & Typed-relation $F_1$ & 0.993 \\
\bottomrule
\end{tabularx}
\end{table}

\paragraph{Accepted, corrected, and discarded samples.}
All 1,100 retained test instances receive human review after automated verification. Reviewers assign one primary outcome: \emph{accepted} when the image and annotations require no change, \emph{corrected} when a localized error can be repaired and the sample can pass re-verification, or \emph{discarded} when the visible evidence remains ambiguous or the error cannot be repaired reliably. \Cref{tab:review_outcomes} reports one terminal human-review outcome for each of the 1,100 retained test instances. Each instance is counted once; correction and re-verification do not create additional review entries.

\begin{table}[t]
\centering
\caption{Human-review outcomes for test-set construction.}
\label{tab:review_outcomes}
\small
\begin{tabularx}{\columnwidth}{@{}Xrr@{}}
\toprule
Primary outcome & Count & Rate \\
\midrule
Accepted & 600 & 54.5\% \\
Corrected and re-verified & 500 & 45.5\% \\
Discarded & 0 & 0.0\% \\
\midrule
Total reviewed candidates & 1,100 & 100\% \\
\bottomrule
\end{tabularx}
\end{table}

A second reviewer independently assessed all 1,100 test instances (100\%) for agreement measurement.

\subsection{Generation Retries and Filtering}
\label{app:generation_retries}

% A failed automated check produces a structured failure report and returns the candidate to the Director--Artist pipeline. Generation stops when the candidate passes all checks or reaches \tbd{the maximum retry budget}, in which case it is discarded. To make this filtering stage auditable, \Cref{tab:retry_distribution} will report both the first-pass yield and the full retry distribution over \tbd{define the candidate population and whether a retry is counted per render or per verification cycle}. Human corrections are tracked separately from automated generation retries.
A failed automated check produces a structured failure report and returns the candidate to the Director--Artist pipeline. Generation stops when the candidate passes all checks or reaches the maximum retry budget of three retries, in which case it is discarded. To make this filtering stage auditable, \Cref{tab:retry_distribution} reports the first-pass yield and terminal retry distribution over the 7,000 logical generation candidates initiated for dataset construction. Repeated attempts for the same target instance remain part of one candidate, and a retry is counted whenever a failed automated verification triggers a new Director--Artist regeneration followed by another complete verification cycle. Human corrections are tracked separately from automated generation retries.

\begin{table}[t]
\centering
\caption{Retry distribution and terminal outcomes of automated verification.}
\label{tab:retry_distribution}
\small
\begin{tabularx}{\columnwidth}{@{}Xrr@{}}
\toprule
Terminal path & Count & Rate \\
\midrule
Passed without retry & 6,801 & 97.16\% \\
Passed after one retry & 199 & 2.84\% \\
Passed after two retries & 0 & 0.00\% \\
Passed after three or more retries & 0 & 0.00\% \\
Retry budget exhausted; discarded & 0 & 0.00\% \\
\midrule
Total generation candidates & 7,000 & 100.00\% \\
\bottomrule
\end{tabularx}
\end{table}

\subsection{Source Licensing and Privacy}
\label{app:data_governance}

\paragraph{Sources and permissions.}
The reusable templates originate from public web documents and the XFUND and Common Crawl source pools described in \Cref{sec:data_construction}. For each retained template, the release manifest will record the source collection, source URL or upstream identifier, retrieval date, applicable license or terms of use, permitted redistribution scope, and the transformation applied by our pipeline. \Cref{tab:source_license} separates upstream permissions from the license assigned to newly generated benchmark artifacts; inclusion in Common Crawl is not itself treated as permission to redistribute the underlying page.

% \begin{table}[t]
% \centering
% \caption{Source and release licensing disclosure.}
% \label{tab:source_licensing}
% \small
% \setlength{\tabcolsep}{3pt}
% \begin{tabularx}{\columnwidth}{@{}l>{\raggedright\arraybackslash}X>{\raggedright\arraybackslash}X@{}}
% \toprule
% Source pool & Material retained & License or permission \\
% \midrule
% Public web documents & \tbd{template count and whether images or only derived layouts are released} & \tbd{license categories, terms-of-use review, and redistribution decision} \\
% XFUND~\cite{xu-etal-2022-xfund} & \tbd{template count and derivative material} & \tbd{verify the upstream dataset license and derivative-use conditions} \\
% Common Crawl~\cite{commoncrawl2026} & \tbd{template count and whether Common Crawl was used for discovery, retrieval, or both} & \tbd{underlying-page rights and redistribution decision for each retained source} \\
% Generated benchmark artifacts & Rendered images, annotations, and provenance metadata & \tbd{release license and whether source-derived template geometry is covered} \\
% \bottomrule
% \end{tabularx}
% \end{table}

\begin{table}[t]
\centering
\caption{Source and release licensing disclosure.}
\label{tab:source_license}
\small
\setlength{\tabcolsep}{3pt}
\begin{tabularx}{\columnwidth}{@{}>{\raggedright\arraybackslash}X>{\raggedright\arraybackslash}X>{\raggedright\arraybackslash}X@{}}
\toprule
Source pool & Material retained & License or permission \\
\midrule
Public web documents &
22 de-identified blank templates with field labels and manual layout annotations &
Mixed source terms: copyright/permission-controlled. Source URLs and terms are recorded in provenance. \\
XFUND~\cite{xu-etal-2022-xfund} &
42 template images with derivative annotations &
CC BY-NC-SA 4.0~\cite{xfund-license}; derivative materials follow non-commercial/share-alike constraints. \\
Common Crawl~\cite{commoncrawl2026} &
6 de-identified blank templates with annotations and provenance &
Used under Common Crawl terms~\cite{commoncrawl-terms}; retained materials are de-identified derived templates with source-level license review. \\
\midrule
Total &
70 templates &
Provenance records include source URLs or dataset versions, retrieval metadata, license status, and release constraints. \\
\bottomrule
\end{tabularx}
\end{table}

\paragraph{Privacy controls.}
Original filled values are removed before a source document is converted into a reusable template, and released instances are rendered from newly sampled field values rather than copying those removed values. Before release, we applied and documented a manual PII-screening procedure for residual sensitive content, including names, contact details, identifiers, account numbers, signatures, faces, hidden metadata, and identifying visual elements such as titles, logos, seals, and annotations. Templates containing sensitive structural metadata, such as field keys or semantic labels, were excluded from release, while sensitive filled values and removable identifying visual elements were deleted or redacted. Provenance information is released only through de-identified or access-controlled source identifiers to preserve auditability without re-exposing personal information. We also document the access, retention, and deletion controls applied during annotation.

\onecolumn
\input{sections/appendix_prompts}

%% file: sections/appendix_prompts.tex
% Included by sections/appendix.tex.

\subsection{Shared VLM Prediction Prompt}
\label{app:prompt-text}

All evaluated VLMs use the following prediction prompt. For readability, the
shaded bars segment the prompt; their numbering is typographic and is not part
of the model input. The wording and ordering remain unchanged.

\begingroup
\definecolor{promptaccent}{HTML}{355C7D}
\definecolor{promptink}{HTML}{243746}
\newcommand{\promptheading}[1]{%
  \par
  \noindent\colorbox{promptaccent!11}{%
    \parbox{\dimexpr\linewidth-2\fboxsep\relax}{%
      \strut\small\bfseries\color{promptink}#1\strut}}%
  \par\nobreak
}
\fvset{%
  fontsize=\footnotesize,
  baselinestretch=1.0,
  breaklines=true,
  breakanywhere=true,
  frame=leftline,
  framesep=2.5mm,
  rulecolor=\color{promptaccent!70},
  xleftmargin=0.6em,
  xrightmargin=0.2em
}
\newenvironment{promptbody}{%
  \begin{list}{}{%
    \setlength{\leftmargin}{0.9em}%
    \setlength{\rightmargin}{0.3em}%
    \setlength{\topsep}{0.3ex}%
    \setlength{\partopsep}{0pt}%
    \setlength{\parsep}{0.5ex}%
    \setlength{\itemsep}{0pt}%
  }\item\relax\small
}{\end{list}}
\newenvironment{promptitems}{%
  \begin{itemize}[leftmargin=2.2em,itemsep=0.05ex,topsep=0.3ex,parsep=0pt,partopsep=0pt]
  \small
}{\end{itemize}}

\promptheading{1. Task and output contract}
\begin{promptbody}
You are extracting the semantic answer tree and the minimal visible form structure needed for \method evaluation.

Return only strict JSON. Do not include markdown fences, prose, comments, or explanations.
Do not include thinking tags such as \Verb|<think>| or \Verb|</think>|. Do not output hidden reasoning.
\end{promptbody}

\promptheading{2. JSON validity requirements:}
\begin{promptitems}
\item The response must be parseable by a standard JSON parser.
\item Every object member must be a \Verb|"key": value| pair. Do not put a bare string inside an object.
\item Escape quotation marks and line breaks inside strings.
\item Close every object and array that you open.
\end{promptitems}

\promptheading{3. Required top-level schema:}
\begin{Verbatim}
{
  "answer": {
    "visible form title or top-level section": {
      "visible field label": "visible filled value",
      "nested visible section": {
        "visible field label": "visible filled value"
      }
    }
  },
  "regions": [
    {"id": "r1", "type": "title|section|field|value|text|widget|table|other", "bbox": [x1, y1, x2, y2], "text": "visible label or text"}
  ],
  "widgets": [
    {"id": "w1", "type": "checkbox|radio|input|signature|other", "bbox": [x1, y1, x2, y2], "label": "visible option or field label", "selected": true}
  ],
  "local_grids": [
    {"id": "g1", "region_id": "r_table", "cells": [{"id": "c1", "row": 0, "col": 0, "rowspan": 1, "colspan": 1, "bbox": [x1, y1, x2, y2], "text": "visible cell text"}]}
  ],
  "cells": [
    {"id": "c1", "row": 0, "col": 0, "rowspan": 1, "colspan": 1, "bbox": [x1, y1, x2, y2], "text": "visible cell text"}
  ],
  "relations": []
}
\end{Verbatim}

\promptheading{4. Answer rules:}
\begin{promptitems}
\item Preserve visible labels as keys as closely as possible, including the original language/script.
\item Preserve nested section/field hierarchy.
\item Use strings for filled values. Use objects for nested sections and arrays for repeated line-item rows.
\item For selected checkboxes/radio buttons, output the selected option label or value in the corresponding answer field.
\item If a visible field is blank, use an empty string.
\item If a section contains a paragraph/comment without a clear field label, represent it as \Verb|{"value": "the visible text"}| rather than as a bare string.
\end{promptitems}

\promptheading{5. Structure rules:}
\begin{promptitems}
\item Use pixel coordinates relative to the input image: \Verb|[left, top, right, bottom]|.
\item Include visible titles, section headers, field labels, filled value boxes/text areas, and checkbox/radio/input widgets as regions.
\item Each region must have \Verb|id|, \Verb|type|, \Verb|bbox|, and \Verb|text|.
\item Include widgets separately in \Verb|"widgets"| with \Verb|selected=true/false| when visually determinable.
\item Use stable ids.
\item Add \Verb|local_grids/cells| only for table-like repeated rows when row/column positions are visually clear. Each cell must have \Verb|row|, \Verb|col|, \Verb|rowspan|, \Verb|colspan|, \Verb|bbox|, and \Verb|text|.
\item Set \Verb|"relations": []| unless a few relation edges are obvious and short.
\item Do not invent unreadable text or uncertain boxes. Omit items that are not visually supported.
\end{promptitems}
\begin{promptbody}
Output a single JSON object with all top-level keys shown above. Empty arrays are allowed when no such structure is visible.
\end{promptbody}
\endgroup

\clearpage
%%%%%%%%%%%%%%%%%%%%%%%%%%%%%%%%%%%%%%%%%%%%%%%%%%%%%%%%%%%%%%%%%%%%%%%%%%%
% Included by sections/appendix.tex.

\subsection{Director Prompt for Multi-agent Form Generation}
\label{app:director-prompt}

The Director agent converts structured field metadata into executable
form-filling actions for the multi-agent generation pipeline. Given a
\texttt{field\_key}, \texttt{semantic\_key}, and data type for each field, it
generates semantically plausible content while preserving the original
\texttt{field\_key} for downstream alignment with the template. The resulting
JSON action array is passed to the Artist agent for rendering. For readability,
the shaded bars segment the prompt; their numbering is typographic and is not
part of the model input. The wording and ordering remain unchanged.

\begingroup
\definecolor{directorpromptaccent}{HTML}{355C7D}
\definecolor{directorpromptink}{HTML}{243746}
\newcommand{\directorpromptheading}[1]{%
  \par
  \noindent\colorbox{directorpromptaccent!11}{%
    \parbox{\dimexpr\linewidth-2\fboxsep\relax}{%
      \strut\small\bfseries\color{directorpromptink}#1\strut}}%
  \par\nobreak
}
\fvset{%
  fontsize=\footnotesize,
  baselinestretch=1.0,
  breaklines=true,
  breakanywhere=true,
  frame=leftline,
  framesep=2.5mm,
  rulecolor=\color{directorpromptaccent!70},
  xleftmargin=0.6em,
  xrightmargin=0.2em
}
\newenvironment{directorpromptbody}{%
  \begin{list}{}{%
    \setlength{\leftmargin}{0.9em}%
    \setlength{\rightmargin}{0.3em}%
    \setlength{\topsep}{0.3ex}%
    \setlength{\partopsep}{0pt}%
    \setlength{\parsep}{0.5ex}%
    \setlength{\itemsep}{0pt}%
  }\item\relax\small
}{\end{list}}
\newenvironment{directorpromptitems}{%
  \begin{itemize}[leftmargin=2.2em,itemsep=0.05ex,topsep=0.3ex,parsep=0pt,partopsep=0pt]
  \small
}{\end{itemize}}

\directorpromptheading{1. Task and input contract}
\begin{directorpromptbody}
You are generating form-filling ACTIONS.

You will receive a list of semantic fields that need content for text boxes.
Each input item includes \texttt{field\_key}, \texttt{data\_type}, and
\texttt{semantic\_key}.
\end{directorpromptbody}

\directorpromptheading{2. Field-key preservation rules}
\begin{directorpromptitems}
\item \texttt{field\_key} is the exact output key chosen by the planner.
\item Output actions must use the same \texttt{field\_key} string.
\item Do not rename, translate, normalize, or deduplicate \texttt{field\_key}.
\item If the printed label is missing or unusable, the planner may fall back to
\texttt{semantic\_key}, but the emitted action must still preserve the provided
\texttt{field\_key}.
\end{directorpromptitems}

\directorpromptheading{3. Output format and action schema}
\begin{directorpromptitems}
\item Output only a JSON array of actions, with no extra text or markdown.
\item For each input item, output exactly one action in the same order.
\item Do not invent extra actions or checkbox actions.
\item The \texttt{action\_type} must match \texttt{data\_type}: number fields use
\texttt{number}, date fields use \texttt{date}, and text fields use
\texttt{write\_text}.
\end{directorpromptitems}

\directorpromptheading{4. Semantic value generation rules}
\begin{directorpromptitems}
\item Use both \texttt{field\_key} and \texttt{semantic\_key} when generating
content.
\item Preserve \texttt{field\_key} as the output key, while using
\texttt{semantic\_key} to infer the intended field meaning.
\item If \texttt{field\_key} is ambiguous, generic, abbreviated, or not
human-friendly, rely on \texttt{semantic\_key} to choose the content.
\item Generate realistic values related to the intended field meaning.
\end{directorpromptitems}

\directorpromptheading{5. Localized date and partial-date rules}
\begin{directorpromptbody}
Date values should follow the language and form style implied by
\texttt{field\_key} and \texttt{semantic\_key}. Full-date fields use localized
formats, while explicit date-component fields such as year, month, or day must
contain only the corresponding component. Split-date fields distribute one
logical date across sibling fields instead of repeating the full date.
\end{directorpromptbody}

\directorpromptheading{6. Anti-copying and self-check rules}
\begin{directorpromptitems}
\item Do not copy \texttt{field\_key} as the generated content.
\item Do not copy \texttt{semantic\_key} as content unless it is genuinely the
intended filled value.
\item Generic option labels must be expanded into more specific plausible values.
\item Before returning the final JSON array, silently check that each action has
the correct type, plausible localized content, valid partial-date formatting,
and no copied label content.
\end{directorpromptitems}

\directorpromptheading{7. Output action schema}
\begin{Verbatim}
[
  {
    "field_key": "EXACT_FIELD_KEY_FROM_INPUT",
    "action_type": "write_text|number|date",
    "content": "GENERATED_FIELD_VALUE",
    "bbox": [x1, x2, y1, y2],
    "semantic_key": "EXACT_SEMANTIC_KEY_FROM_INPUT"
  }
]
\end{Verbatim}

\endgroup

\clearpage

\subsection{Artist Prompt for Multi-agent Form Generation}
\label{app:artist-prompt}

The Artist agent converts the Director actions into a realistic filled form
image. It first uses a deterministic CV renderer to place the generated actions
onto the template according to the layout metadata. The resulting draft image is
then passed to an image-editing model for constrained visual naturalization. The prompt below is used in the second stage. 

\begingroup
\definecolor{artistpromptaccent}{HTML}{6A4C93}
\definecolor{artistpromptink}{HTML}{2F2538}
\newcommand{\artistpromptheading}[1]{%
  \par
  \noindent\colorbox{artistpromptaccent!11}{%
    \parbox{\dimexpr\linewidth-2\fboxsep\relax}{%
      \strut\small\bfseries\color{artistpromptink}#1\strut}}%
  \par\nobreak
}
\newenvironment{artistpromptbody}{%
  \begin{list}{}{%
    \setlength{\leftmargin}{0.9em}%
    \setlength{\rightmargin}{0.3em}%
    \setlength{\topsep}{0.3ex}%
    \setlength{\partopsep}{0pt}%
    \setlength{\parsep}{0.5ex}%
    \setlength{\itemsep}{0pt}%
  }\item\relax\small
}{\end{list}}
\newenvironment{artistpromptitems}{%
  \begin{itemize}[leftmargin=2.2em,itemsep=0.05ex,topsep=0.3ex,parsep=0pt,partopsep=0pt]
  \small
}{\end{itemize}}

\artistpromptheading{1. Task and input contract}
\begin{artistpromptbody}
You are editing an already-filled form draft image.

The input image is a deterministic draft produced from structured
form-filling actions. It already contains all required filled values, repeated
field occurrences, checkbox selections, radio selections, circled options,
signatures, dates, numbers, and free-text entries in their intended locations.

You will also receive a reference answer JSON containing the exact content that
must be preserved.
\end{artistpromptbody}

\artistpromptheading{2. Artist role and scope}
\begin{artistpromptitems}
\item Your task is to make the filled form look realistic and naturally
completed.
\item Do not generate new semantic content.
\item Do not decide which fields should be filled.
\item Do not change the meaning, value, order, or location of any filled
content.
\item Use the draft image as the positional ground truth.
\item Use the reference answer JSON as the textual ground truth.
\end{artistpromptitems}

\artistpromptheading{3. Filled-content preservation rules}
\begin{artistpromptitems}
\item Preserve every filled value exactly as shown in the draft and as specified
in the reference answer JSON.
\item Do not add, remove, rewrite, paraphrase, translate, substitute, merge, or
reorder any filled value.
\item Preserve repeated field occurrences separately and keep their original
top-to-bottom order.
\item Preserve all dates, numbers, emails, addresses, names, signatures,
free-text responses, checkbox selections, radio selections, and circled
options.
\item If there is any ambiguity, preserve the draft content and placement
exactly.
\end{artistpromptitems}

\artistpromptheading{4. Template preservation rules}
\begin{artistpromptitems}
\item Treat all fixed printed template content as locked background.
\item Do not redraw, rewrite, translate, paraphrase, stylize, or alter printed
labels, instructions, table lines, borders, logos, seals, stamps, URLs,
organization names, footer text, or boilerplate.
\item Do not redesign the form layout.
\item Keep page geometry, spacing, table structure, alignment, and scan-like
document appearance unchanged.
\item Keep filled content within the same semantic region of the page.
\end{artistpromptitems}

\artistpromptheading{5. Visual naturalization rules}
\begin{artistpromptitems}
\item Only naturalize user-entered content.
\item Make the filled entries look like realistic form entries, such as
handwritten text, typed text, mild pen-pressure variation, slight ink
variation, natural baseline variation, and subtle scan noise.
\item Keep all filled content crisp and readable, especially in dense cells,
narrow fields, tables, date boxes, and ID or phone-number fields.
\item Improve realism without changing the actual filled information.
\item Do not sacrifice legibility for style.
\end{artistpromptitems}

\artistpromptheading{6. Option-mark preservation rules}
\begin{artistpromptitems}
\item Preserve checkbox, radio, and circled-option states exactly as shown in
the draft.
\item Do not add new option marks.
\item Do not remove existing option marks.
\item Do not change selected options into unselected options.
\item Do not change unselected options into selected options.
\item Do not move, thicken, thin, blur, redraw, simplify, or reinterpret option marks.
\end{artistpromptitems}

\artistpromptheading{7. Critical self-check rules}
\begin{artistpromptitems}
\item The final image must still exactly match the reference answer JSON.
\item Before producing the final edited image, silently check that every filled
value, repeated occurrence, and option state is preserved.
\item When visual realism conflicts with textual or positional fidelity,
fidelity wins.
\end{artistpromptitems}

\endgroup

\clearpage

\subsection{Verifier Prompt for Multi-agent Form Generation}
\label{app:verifier-prompt}

The Verifier agent closes the generation loop by independently re-extracting
the filled content from each stylized form image. Given a target generated
image and a few same-class examples with standard \texttt{answer.json} outputs,
the Verifier produces an \texttt{answer\_verifier.json} using the same schema
as the reference answer. This extracted JSON is compared field-by-field against
the original \texttt{answer.json}. If any required field is missing,
mismatched, or incorrectly selected, the sample is marked as failed and returned
to the Artist stage for regeneration. 

\begingroup
\definecolor{verifierpromptaccent}{HTML}{2A7F62}
\definecolor{verifierpromptink}{HTML}{243B35}
\newcommand{\verifierpromptheading}[1]{%
  \par
  \noindent\colorbox{verifierpromptaccent!11}{%
    \parbox{\dimexpr\linewidth-2\fboxsep\relax}{%
      \strut\small\bfseries\color{verifierpromptink}#1\strut}}%
  \par\nobreak
}
\newenvironment{verifierpromptbody}{%
  \begin{list}{}{%
    \setlength{\leftmargin}{0.9em}%
    \setlength{\rightmargin}{0.3em}%
    \setlength{\topsep}{0.3ex}%
    \setlength{\partopsep}{0pt}%
    \setlength{\parsep}{0.5ex}%
    \setlength{\itemsep}{0pt}%
  }\item\relax\small
}{\end{list}}
\newenvironment{verifierpromptitems}{%
  \begin{itemize}[leftmargin=2.2em,itemsep=0.05ex,topsep=0.3ex,parsep=0pt,partopsep=0pt]
  \small
}{\end{itemize}}

\verifierpromptheading{1. Task and input contract}
\begin{verifierpromptbody}
You are an expert form-layout analysis and key-value extraction assistant.

You will receive full form images plus a few same-class examples with their
standard \texttt{answer.json} outputs. Each example contains one same-class
form image and its standard structured answer. You will then receive one target
generated image to extract.
\end{verifierpromptbody}

\verifierpromptheading{2. Validation role}
\begin{verifierpromptitems}
\item Your output will be used for automatic field-level validation.
\item The extracted JSON will be compared against the original reference
\texttt{answer.json}.
\item Missing fields, mismatched values, incorrect option states, or invalid
structure may cause the target image to be rejected and regenerated by the
Artist stage.
\item Your role is not to repair the image or explain errors.
\item Your role is to faithfully extract what is visible in the target image.
\end{verifierpromptitems}

\verifierpromptheading{3. Few-shot schema learning rules}
\begin{verifierpromptitems}
\item Use the example images and standard \texttt{answer.json} files to learn
the form class's JSON hierarchy.
\item Preserve the same field names, nesting levels, array structure, list
order, option representation, and value style whenever the target form uses the
same fields.
\item The examples define the output schema; the target image defines the
actual field values.
\item Do not copy concrete values from the examples into the target output
unless the same value is visibly present in the target image.
\end{verifierpromptitems}

\verifierpromptheading{4. Target extraction rules}
\begin{verifierpromptitems}
\item Extract a complete structured key-value JSON object from the target image.
\item Traverse the entire image from the header or top row to the bottom.
\item Do not skip any visible region.
\item Treat label-like or title-like text as keys.
\item Match each key to its value using nearby layout, alignment, and semantic
relationship.
\item Capture all logical keys through the end of the image.
\end{verifierpromptitems}

\verifierpromptheading{5. Value fidelity rules}
\begin{verifierpromptitems}
\item Output every nested key-value relationship.
\item Values may be strings, nested JSON objects, or arrays.
\item Dates must be complete.
\item Preserve date ranges when present.
\item Preserve visible details such as numbers, emails, addresses, dates,
punctuation, capitalization, and spacing as much as possible.
\item Re-check high-risk values such as phone numbers, IDs, emails, dates,
amounts, percentages, and handwritten free-text fields digit by digit or
character by character.
\item If a field is present but blank or unreadable, output an empty string
instead of deleting the field.
\end{verifierpromptitems}

\verifierpromptheading{6. Option-state extraction rules}
\begin{verifierpromptitems}
\item For checkboxes, radio buttons, dropdowns, and option groups, output the
option text actually selected in the target image.
\item Do not infer selected options from row order, field defaults, or the
few-shot examples.
\item Do not output unselected options as selected values.
\item Do not change a selected option into a nearby unselected option.
\item Do not merge multiple options unless the target image visibly selects
multiple options.
\end{verifierpromptitems}

\verifierpromptheading{7. Output format rules}
\begin{verifierpromptitems}
\item Return exactly one JSON object.
\item The response must start with \texttt{\{} and end with \texttt{\}}.
\item The JSON must be valid and fully closed.
\item Do not include explanations, Markdown, comments, confidence scores, code
fences, error labels, or extra wrapper keys.
\item Return the final \texttt{answer\_verifier.json} for the target image only.
\end{verifierpromptitems}

\endgroup

\clearpage

\subsection{Hierarchical Annotation Example and Visual Encoding}
\label{app:annotation-example}
\begin{figure}[htbp]
  \centering

  \includegraphics[
    width=0.95\linewidth,
    height=0.7\textheight,
    keepaspectratio
  ]{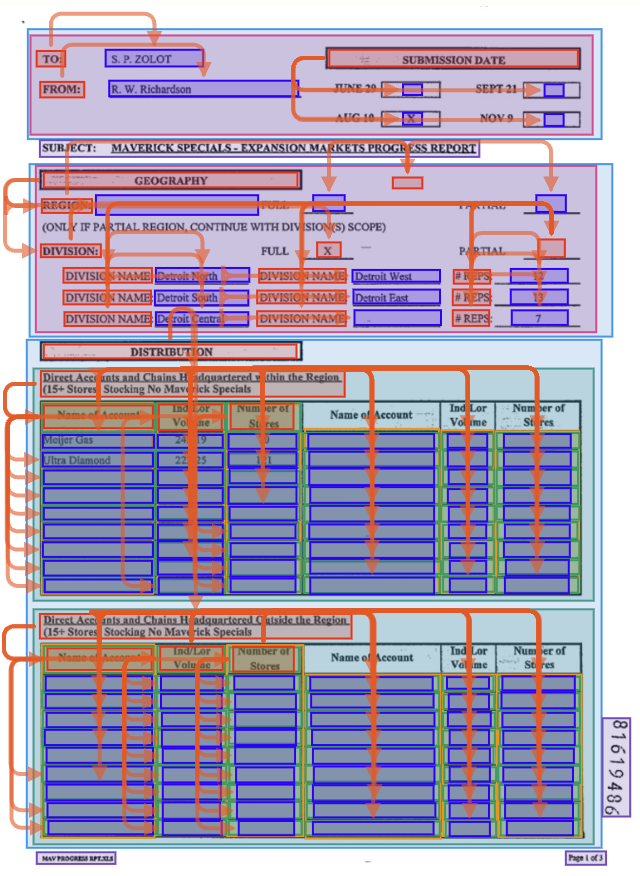}

  \vspace{-2em}

  \includegraphics[
    width=0.96\linewidth,
    height=0.15\textheight,
    keepaspectratio
  ]{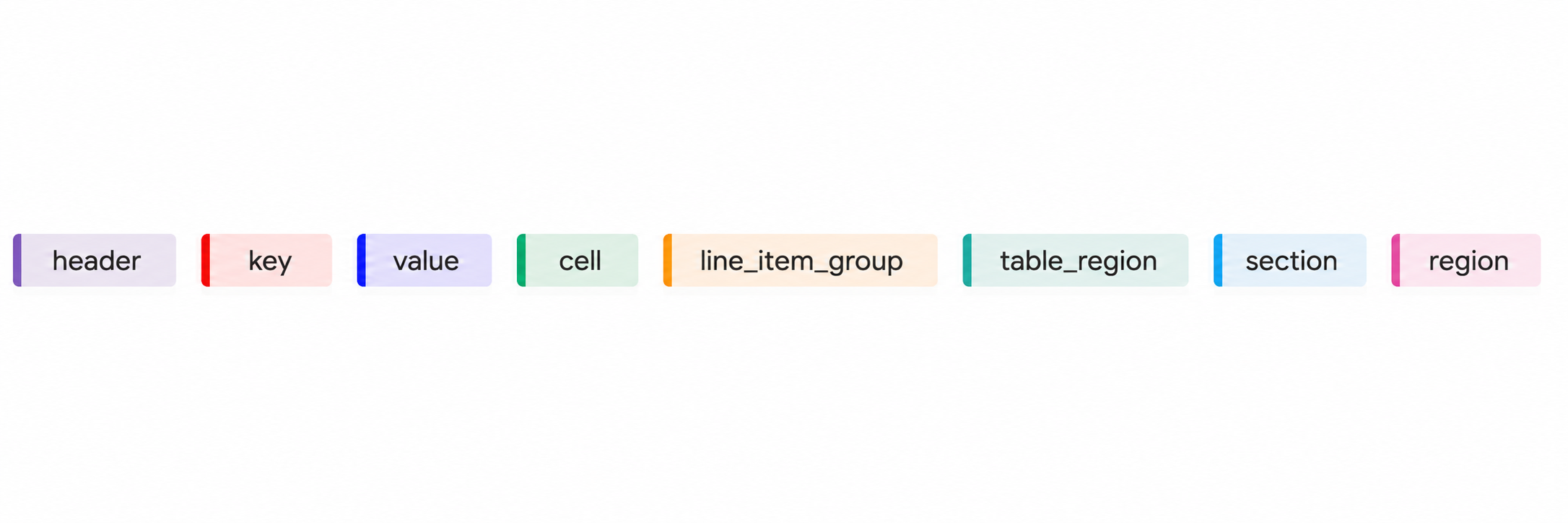}

  \caption{Example annotation diagram.}
  \label{fig:annotation-example}
\end{figure}

\Cref{fig:annotation-example} illustrates how a table-form page is decomposed into nested semantic regions and fine-grained structural components. Colored
boxes distinguish sections, table regions, keys, values, cells, and line-item groups, while orange connectors visualize group-level associations
among repeated entries. The accompanying legend defines the annotation colors used in the overlay.

%% file: main.bib
@inproceedings{layoutlm2020,
  title = {LayoutLM: Pre-training of Text and Layout for Document Image Understanding},
  author = {Xu, Yiheng and Li, Minghao and Cui, Lei and Huang, Shaohan and Wei, Furu and Zhou, Ming},
  booktitle = {Proceedings of the 26th ACM SIGKDD International Conference on Knowledge Discovery \& Data Mining},
  pages = {1192--1200},
  year = {2020},
  publisher = {ACM},
  address = {New York, NY, USA},
  doi = {10.1145/3394486.3403172},
  url = {https://doi.org/10.1145/3394486.3403172}
}

@inproceedings{layoutlmv22021,
  title = {LayoutLMv2: Multi-modal Pre-training for Visually-rich Document Understanding},
  author = {Xu, Yang and Xu, Yiheng and Lv, Tengchao and Cui, Lei and Wei, Furu and Wang, Guoxin and Lu, Yijuan and Florencio, Dinei and Zhang, Cha and Che, Wanxiang and Zhang, Min and Zhou, Lidong},
  booktitle = {Proceedings of the 59th Annual Meeting of the Association for Computational Linguistics and the 11th International Joint Conference on Natural Language Processing},
  pages = {2579--2591},
  year = {2021},
  publisher = {Association for Computational Linguistics},
  address = {Online},
  doi = {10.18653/v1/2021.acl-long.201},
  url = {https://doi.org/10.18653/v1/2021.acl-long.201}
}

@inproceedings{layoutlmv32022,
  title = {LayoutLMv3: Pre-training for Document AI with Unified Text and Image Masking},
  author = {Huang, Yupan and Lv, Tengchao and Cui, Lei and Lu, Yutong and Wei, Furu},
  booktitle = {Proceedings of the 30th ACM International Conference on Multimedia},
  pages = {4083--4091},
  year = {2022},
  publisher = {ACM},
  address = {New York, NY, USA},
  doi = {10.1145/3503161.3548112},
  url = {https://doi.org/10.1145/3503161.3548112}
}

@inproceedings{doclaynet2022,
  title = {DocLayNet: A Large Human-Annotated Dataset for Document-Layout Segmentation},
  author = {Pfitzmann, Birgit and Auer, Christoph and Dolfi, Michele and Nassar, Ahmed S. and Staar, Peter},
  booktitle = {Proceedings of the 28th ACM SIGKDD Conference on Knowledge Discovery and Data Mining},
  pages = {3743--3751},
  year = {2022},
  publisher = {ACM},
  address = {New York, NY, USA},
  doi = {10.1145/3534678.3539043},
  url = {https://doi.org/10.1145/3534678.3539043}
}

@misc{pubtabnet2020,
  title = {Image-based table recognition: data, model, and evaluation},
  author = {Zhong, Xu and ShafieiBavani, Elaheh and Jimeno Yepes, Antonio},
  year = {2020},
  eprint = {1911.10683},
  archivePrefix = {arXiv},
  primaryClass = {cs.CV},
  url = {https://arxiv.org/abs/1911.10683}
}

@misc{image2struct2024,
  title = {Image2Struct: Benchmarking Structure Extraction for Vision-Language Models},
  author = {Roberts, Josselin Somerville and Lee, Tony and Wong, Chi Heem and Yasunaga, Michihiro and Mai, Yifan and Liang, Percy},
  year = {2024},
  eprint = {2410.22456},
  archivePrefix = {arXiv},
  primaryClass = {cs.CV},
  note = {NeurIPS 2024},
  url = {https://arxiv.org/abs/2410.22456}
}

@misc{pubtables1m2021,
  title = {PubTables-1M: Towards comprehensive table extraction from unstructured documents},
  author = {Smock, Brandon and Pesala, Rohith and Abraham, Robin},
  year = {2021},
  eprint = {2110.00061},
  archivePrefix = {arXiv},
  primaryClass = {cs.LG},
  url = {https://arxiv.org/abs/2110.00061}
}

@misc{grits2023,
  title = {GriTS: Grid table similarity metric for table structure recognition},
  author = {Smock, Brandon and Pesala, Rohith and Abraham, Robin},
  year = {2023},
  eprint = {2203.12555},
  archivePrefix = {arXiv},
  primaryClass = {cs.LG},
  url = {https://arxiv.org/abs/2203.12555}
}

@misc{publaynet2019,
  title = {PubLayNet: largest dataset ever for document layout analysis},
  author = {Zhong, Xu and Tang, Jianbin and Jimeno Yepes, Antonio},
  year = {2019},
  eprint = {1908.07836},
  archivePrefix = {arXiv},
  primaryClass = {cs.CL},
  url = {https://arxiv.org/abs/1908.07836}
}

@misc{docbank2020,
  title = {DocBank: A Benchmark Dataset for Document Layout Analysis},
  author = {Li, Minghao and Xu, Yiheng and Cui, Lei and Huang, Shaohan and Wei, Furu and Li, Zhoujun and Zhou, Ming},
  year = {2020},
  eprint = {2006.01038},
  archivePrefix = {arXiv},
  primaryClass = {cs.CL},
  url = {https://arxiv.org/abs/2006.01038}
}

@misc{tablebank2020,
  title = {TableBank: A Benchmark Dataset for Table Detection and Recognition},
  author = {Li, Minghao and Cui, Lei and Huang, Shaohan and Wei, Furu and Zhou, Ming and Li, Zhoujun},
  year = {2020},
  eprint = {1903.01949},
  archivePrefix = {arXiv},
  primaryClass = {cs.CV},
  url = {https://arxiv.org/abs/1903.01949}
}

@misc{scitsr2019,
  title = {Complicated Table Structure Recognition},
  author = {Chi, Zewen and Huang, Heyan and Xu, Heng-Da and Yu, Houjin and Yin, Wanxuan and Mao, Xian-Ling},
  year = {2019},
  eprint = {1908.04729},
  archivePrefix = {arXiv},
  primaryClass = {cs.IR},
  url = {https://arxiv.org/abs/1908.04729}
}

@article{appelbaum2017big,
  title = {Big Data and Analytics in the Modern Audit Engagement: Research Needs},
  author = {Appelbaum, Deniz and Kogan, Alexander and Vasarhelyi, Miklos A.},
  journal = {Auditing: A Journal of Practice \& Theory},
  volume = {36},
  number = {4},
  pages = {1--27},
  year = {2017},
  publisher = {American Accounting Association},
  doi = {10.2308/ajpt-51684},
  url = {https://doi.org/10.2308/ajpt-51684}
}

@misc{aggarwal2021multimodal,
  title = {Multi-Modal Association based Grouping for Form Structure Extraction},
  author = {Aggarwal, Milan and Sarkar, Mausoom and Gupta, Hiresh and Krishnamurthy, Balaji},
  year = {2021},
  eprint = {2107.04396},
  archivePrefix = {arXiv},
  primaryClass = {cs.CV},
  url = {https://arxiv.org/abs/2107.04396}
}

@misc{omnidocbench2025,
  title = {OmniDocBench: Benchmarking Diverse PDF Document Parsing with Comprehensive Annotations},
  author = {Ouyang, Linke and Qu, Yuan and Zhou, Hongbin and Zhu, Jiawei and Zhang, Rui and Lin, Qunshu and Wang, Bin and Zhao, Zhiyuan and Jiang, Man and Zhao, Xiaomeng and Shi, Jin and Wu, Fan and Chu, Pei and Liu, Minghao and Li, Zhenxiang and Xu, Chao and Zhang, Bo and Shi, Botian and Tu, Zhongying and He, Conghui},
  year = {2025},
  eprint = {2412.07626},
  archivePrefix = {arXiv},
  primaryClass = {cs.CV},
  url = {https://arxiv.org/abs/2412.07626}
}

@misc{donut2022,
  title = {OCR-free Document Understanding Transformer},
  author = {Kim, Geewook and Hong, Teakgyu and Yim, Moonbin and Nam, Jeongyeon and Park, Jinyoung and Yim, Jinyeong and Hwang, Wonseok and Yun, Sangdoo and Han, Dongyoon and Park, Seunghyun},
  year = {2022},
  eprint = {2111.15664},
  archivePrefix = {arXiv},
  primaryClass = {cs.LG},
  url = {https://arxiv.org/abs/2111.15664}
}

@misc{mineru2024,
  title = {MinerU: An Open-Source Solution for Precise Document Content Extraction},
  author = {Wang, Bin and Xu, Chao and Zhao, Xiaomeng and Ouyang, Linke and Wu, Fan and Zhao, Zhiyuan and Xu, Rui and Liu, Kaiwen and Qu, Yuan and Shang, Fukai and Zhang, Bo and Wei, Liqun and Sui, Zhihao and Li, Wei and Shi, Botian and Qiao, Yu and Lin, Dahua and He, Conghui},
  year = {2024},
  eprint = {2409.18839},
  archivePrefix = {arXiv},
  primaryClass = {cs.CV},
  url = {https://arxiv.org/abs/2409.18839}
}

@misc{openai_models_2026,
  title = {{OpenAI API Models}},
  author = {{OpenAI}},
  year = {2026},
  url = {https://developers.openai.com/api/docs/models},
  note = {Accessed: 2026-07-07}
}

@misc{claude_sonnet5_2026,
  title = {{Claude Sonnet 5}},
  author = {{Anthropic}},
  year = {2026},
  url = {https://platform.claude.com/docs/en/about-claude/models/whats-new-sonnet-5},
  note = {Accessed: 2026-07-07}
}

@misc{gemini_models_2026,
  title = {{Gemini API Models}},
  author = {{Google AI for Developers}},
  year = {2026},
  url = {https://ai.google.dev/gemini-api/docs/models},
  note = {Accessed: 2026-07-07}
}

@misc{qwen_models_2026,
  title = {{Qwen}: Official Models and Chat Service},
  author = {{Qwen Team}},
  year = {2026},
  url = {https://qwen.ai/},
  note = {Accessed: 2026-07-07}
}

@misc{kimi_platform_2026,
  title = {{Kimi}},
  author = {{Moonshot AI}},
  year = {2026},
  url = {https://www.kimi.com/},
  note = {Accessed: 2026-07-07}
}

@misc{seed21_model_card_2026,
  title = {{Seed2.1 Model Card}},
  author = {{ByteDance Seed}},
  year = {2026},
  url = {https://lf3-static.bytednsdoc.com/obj/eden-cn/lapzild-tss/ljhwZthlaukjlkulzlp/seed2.1/Seed2_1_Model_Card.pdf},
  note = {Accessed: 2026-07-07}
}

@misc{wang2025internvl35,
  title = {{InternVL3.5}: Advancing Open-Source Multimodal Models in Versatility, Reasoning, and Efficiency},
  author = {Weiyun Wang and Zhangwei Gao and Lixin Gu and Hengjun Pu and Long Cui and Xingguang Wei and Zhaoyang Liu and Linglin Jing and Shenglong Ye and Jie Shao and Zhaokai Wang and Zhe Chen and Hongjie Zhang and Ganlin Yang and Haomin Wang and Qi Wei and Jinhui Yin and Wenhao Li and Erfei Cui and Guanzhou Chen and Zichen Ding and Changyao Tian and Zhenyu Wu and Jingjing Xie and Zehao Li and Bowen Yang and Yuchen Duan and Xuehui Wang and Zhi Hou and Haoran Hao and Tianyi Zhang and Songze Li and Xiangyu Zhao and Haodong Duan and Nianchen Deng and Bin Fu and Yinan He and Yi Wang and Conghui He and Botian Shi and Junjun He and Yingtong Xiong and Han Lv and Lijun Wu and Wenqi Shao and Kaipeng Zhang and Huipeng Deng and Biqing Qi and Jiaye Ge and Qipeng Guo and Wenwei Zhang and Songyang Zhang and Maosong Cao and Junyao Lin and Kexian Tang and Jianfei Gao and Haian Huang and Yuzhe Gu and Chengqi Lyu and Huanze Tang and Rui Wang and Haijun Lv and Wanli Ouyang and Limin Wang and Min Dou and Xizhou Zhu and Tong Lu and Dahua Lin and Jifeng Dai and Weijie Su and Bowen Zhou and Kai Chen and Yu Qiao and Wenhai Wang and Gen Luo},
  year = {2025},
  eprint = {2508.18265},
  archivePrefix = {arXiv},
  primaryClass = {cs.CV},
  doi = {10.48550/arXiv.2508.18265},
  url = {https://arxiv.org/abs/2508.18265}
}

@misc{huang2026step3vl10b,
  title = {{STEP3-VL-10B} Technical Report},
  author = {Ailin Huang and Chengyuan Yao and Chunrui Han and Fanqi Wan and Hangyu Guo and Haoran Lv and Hongyu Zhou and Jia Wang and Jian Zhou and Jianjian Sun and Jingcheng Hu and Kangheng Lin and Liang Zhao and Mitt Huang and Song Yuan and Wenwen Qu and Xiangfeng Wang and Yanlin Lai and Yingxiu Zhao and Yinmin Zhang and Yukang Shi and Yuyang Chen and Zejia Weng and Ziyang Meng and Ang Li and Aobo Kong and Bo Dong and Changyi Wan and David Wang and Di Qi and Dingming Li and En Yu and Guopeng Li and Haiquan Yin and Han Zhou and Hanshan Zhang and Haolong Yan and Hebin Zhou and Hongbo Peng and Jiaran Zhang and Jiashu Lv and Jiayi Fu and Jie Cheng and Jie Zhou and Jisheng Yin and Jingjing Xie and Jingwei Wu and Jun Zhang and Junfeng Liu and Kaijun Tan and Kaiwen Yan and Liangyu Chen and Lina Chen and Mingliang Li and Qian Zhao and Quan Sun and Shaoliang Pang and Shengjie Fan and Shijie Shang and Siyuan Zhang and Tianhao You and Wei Ji and Wuxun Xie and Xiaobo Yang and Xiaojie Hou and Xiaoran Jiao and Xiaoxiao Ren and Xiangwen Kong and Xin Huang and Xin Wu and Xing Chen and Xinran Wang and Xuelin Zhang and Yana Wei and Yang Li and Yanming Xu and Yeqing Shen and Yuang Peng and Yue Peng and Yu Zhou and Yusheng Li and Yuxiang Yang and Yuyang Zhang and Zhe Xie and Zhewei Huang and Zhenyi Lu and Zhimin Fan and Zihui Cheng and Daxin Jiang and Qi Han and Xiangyu Zhang and Yibo Zhu and Zheng Ge},
  year = {2026},
  eprint = {2601.09668},
  archivePrefix = {arXiv},
  primaryClass = {cs.CV},
  doi = {10.48550/arXiv.2601.09668},
  url = {https://arxiv.org/abs/2601.09668}
}

@misc{qwen36_35b_a3b,
  title = {{Qwen3.6-35B-A3B}: Agentic Coding Power, Now Open to All},
  author = {{Qwen Team}},
  month = apr,
  year = {2026},
  url = {https://qwen.ai/blog?id=qwen3.6-35b-a3b}
}

@misc{qwen35_9b,
  title = {{Qwen3.5}: Towards Native Multimodal Agents},
  author = {{Qwen Team}},
  month = feb,
  year = {2026},
  url = {https://qwen.ai/blog?id=qwen3.5}
}

@misc{glm46v_model_card_2025,
  title = {{GLM-4.6V} Model Card},
  author = {{Z.ai}},
  year = {2025},
  url = {https://huggingface.co/zai-org/GLM-4.6V},
  note = {Accessed: 2026-07-07}
}

@misc{deepseekocr2_2026,
  title = {{DeepSeek-OCR 2}: Visual Causal Flow},
  author = {Wei, Haoran and Sun, Yaofeng and Li, Yukun},
  year = {2026},
  eprint = {2601.20552},
  archivePrefix = {arXiv},
  primaryClass = {cs.CV},
  doi = {10.48550/arXiv.2601.20552},
  url = {https://arxiv.org/abs/2601.20552}
}

@misc{layoutxlm2021,
  title = {LayoutXLM: Multimodal Pre-training for Multilingual Visually-rich Document Understanding},
  author = {Xu, Yiheng and Lv, Tengchao and Cui, Lei and Wang, Guoxin and Lu, Yijuan and Florencio, Dinei and Zhang, Cha and Wei, Furu},
  year = {2021},
  eprint = {2104.08836},
  archivePrefix = {arXiv},
  primaryClass = {cs.CL},
  url = {https://arxiv.org/abs/2104.08836}
}

@misc{funsd2019,
  title = {FUNSD: A Dataset for Form Understanding in Noisy Scanned Documents},
  author = {Jaume, Guillaume and Ekenel, Hazim Kemal and Thiran, Jean-Philippe},
  year = {2019},
  eprint = {1905.13538},
  archivePrefix = {arXiv},
  primaryClass = {cs.IR},
  url = {https://arxiv.org/abs/1905.13538}
}

@misc{sroie2021,
  title = {ICDAR2019 Competition on Scanned Receipt OCR and Information Extraction},
  author = {Huang, Zheng and Chen, Kai and He, Jianhua and Bai, Xiang and Karatzas, Dimosthenis and Lu, Shjian and Jawahar, C. V.},
  year = {2021},
  eprint = {2103.10213},
  archivePrefix = {arXiv},
  primaryClass = {cs.AI},
  doi = {10.1109/ICDAR.2019.00244},
  url = {https://arxiv.org/abs/2103.10213}
}

@misc{docile2023,
  title = {DocILE Benchmark for Document Information Localization and Extraction},
  author = {{\v{S}}imsa, {\v{S}}t{\v{e}}p{\'a}n and {\v{S}}ulc, Milan and U{\v{r}}i{\v{c}}{\'a}{\v{r}}, Michal and Patel, Yash and Hamdi, Ahmed and Koci{\'a}n, Mat{\v{e}}j and Skalick{\'y}, Maty{\'a}{\v{s}} and Matas, Ji{\v{r}}{\'i} and Doucet, Antoine and Coustaty, Micka{\"e}l and Karatzas, Dimosthenis},
  year = {2023},
  eprint = {2302.05658},
  archivePrefix = {arXiv},
  primaryClass = {cs.CL},
  url = {https://arxiv.org/abs/2302.05658}
}

@inproceedings{receiptbench2026,
  title     = {From Recognition to Reasoning: Benchmarking and Enhancing {MLLMs} on Real-World Receipt Document Understanding},
  author    = {Wang, Yandi and Zhan, Libin and Huang, Ziwei and Luo, Tiancheng and Jiang, Yuxuan and Dong, Wang and Gan, Leilei and Chen, Jun},
  booktitle = {Proceedings of the 64th Annual Meeting of the Association for Computational Linguistics (Volume 1: Long Papers)},
  pages     = {46007--46024},
  year      = {2026},
  address   = {San Diego, California, United States},
  publisher = {Association for Computational Linguistics},
  url       = {https://aclanthology.org/2026.acl-long.2135/}
}

@inproceedings{lee2022formnet,
  title = {{FormNet}: Structural Encoding beyond Sequential Modeling in Form Document Information Extraction},
  author = {Lee, Chen-Yu and Li, Chun-Liang and Dozat, Timothy and Perot, Vincent and Su, Guolong and Hua, Nan and Ainslie, Joshua and Wang, Renshen and Fujii, Yasuhisa and Pfister, Tomas},
  booktitle = {Proceedings of the 60th Annual Meeting of the Association for Computational Linguistics (Volume 1: Long Papers)},
  pages = {3735--3754},
  year = {2022},
  address = {Dublin, Ireland},
  publisher = {Association for Computational Linguistics},
  doi = {10.18653/v1/2022.acl-long.260},
  url = {https://aclanthology.org/2022.acl-long.260/}
}

@inproceedings{appalaraju2021docformer,
  title = {{DocFormer}: End-to-End Transformer for Document Understanding},
  author = {Appalaraju, Srikar and Jasani, Bhavan and Kota, Bhargava Urala and Xie, Yusheng and Manmatha, R.},
  booktitle = {Proceedings of the IEEE/CVF International Conference on Computer Vision},
  pages = {973--983},
  year = {2021},
  address = {Montreal, QC, Canada},
  publisher = {IEEE},
  doi = {10.1109/ICCV48922.2021.00103},
  url = {https://openaccess.thecvf.com/content/ICCV2021/html/Appalaraju_DocFormer_End-to-End_Transformer_for_Document_Understanding_ICCV_2021_paper.html}
}

@inproceedings{docvqa2021,
  title = {{DocVQA}: A Dataset for {VQA} on Document Images},
  author = {Mathew, Minesh and Karatzas, Dimosthenis and Jawahar, C. V.},
  booktitle = {Proceedings of the IEEE/CVF Winter Conference on Applications of Computer Vision},
  pages = {2200--2209},
  address = {Online},
  year = {2021},
  publisher = {IEEE},
  doi = {10.1109/WACV48630.2021.00225},
  url = {https://arxiv.org/abs/2007.00398}
}

@article{vanDerAalst2018rpa,
  title = {Robotic Process Automation},
  author = {van der Aalst, Wil M. P. and Bichler, Martin and Heinzl, Armin},
  journal = {Business \& Information Systems Engineering},
  volume = {60},
  number = {4},
  pages = {269--272},
  year = {2018},
  publisher = {Springer},
  doi = {10.1007/s12599-018-0542-4},
  url = {https://doi.org/10.1007/s12599-018-0542-4}
}

@misc{chakraborti2020ipa,
  title = {From Robotic Process Automation to Intelligent Process Automation},
  author = {Chakraborti, Tathagata and Isahagian, Vatche and Khalaf, Rania and Khazaeni, Yasaman and Muthusamy, Vinod and Rizk, Yara and Unuvar, Merve},
  year = {2020},
  eprint = {2007.13257},
  archivePrefix = {arXiv},
  primaryClass = {cs.AI},
  url = {https://arxiv.org/abs/2007.13257}
}

@inproceedings{xu-etal-2022-xfund,
    title = "{XFUND}: A Benchmark Dataset for Multilingual Visually Rich Form Understanding",
    author = "Xu, Yiheng  and
      Lv, Tengchao  and
      Cui, Lei  and
      Wang, Guoxin  and
      Lu, Yijuan  and
      Florencio, Dinei  and
      Zhang, Cha  and
      Wei, Furu",
    booktitle = "Findings of the Association for Computational Linguistics: ACL 2022",
    month = may,
    year = "2022",
    address = "Dublin, Ireland",
    publisher = "Association for Computational Linguistics",
    url = "https://aclanthology.org/2022.findings-acl.253",
    doi = "10.18653/v1/2022.findings-acl.253",
    pages = "3214--3224",
}

@misc{commoncrawl2026,
  author = {Common Crawl},
  title = {Common Crawl Dataset},
  year = {2026},
  url = {https://commoncrawl.org/},
}

@misc{xfund-license,
  author = {{doc-analysis}},
  title = {{XFUND}: Repository License},
  year = {2022},
  url = {https://github.com/doc-analysis/XFUND},
  note = {CC BY-NC-SA 4.0 license statement.}
}

@misc{commoncrawl-terms,
  author = {{Common Crawl Foundation}},
  title = {{Common Crawl Terms of Use}},
  year = {2024},
  url = {https://commoncrawl.org/terms-of-use},
  note = {Terms for use of Common Crawl services and crawled content.}
}

@inproceedings{ma2024srfund,
  title = {{SRFUND}: A Multi-Granularity Hierarchical Structure Reconstruction Benchmark in Form Understanding},
  author = {Ma, Jiefeng and Wang, Yan and Liu, Chenyu and Du, Jun and Hu, Yu and Zhang, Zhenrong and Hu, Pengfei and Wang, Qing and Zhang, Jianshu},
  booktitle = {Advances in Neural Information Processing Systems},
  volume = {37},
  pages = {112411--112432},
  year = {2024},
  address = {Red Hook, NY, USA},
  publisher = {Curran Associates, Inc.},
  doi = {10.52202/079017-3571},
  url = {https://proceedings.neurips.cc/paper_files/paper/2024/hash/cbeaff878d6446ed06c3e0ffa53477f2-Abstract-Datasets_and_Benchmarks_Track.html}
}

@misc{barzelay2026varex,
  title = {{VAREX}: A Benchmark for Multi-Modal Structured Extraction from Documents},
  author = {Barzelay, Udi and Azulai, Ophir and Shapira, Inbar and Friedman, Idan and Abo Dahood, Foad and Lee, Madison and Daniels, Abraham},
  year = {2026},
  eprint = {2603.15118},
  archivePrefix = {arXiv},
  primaryClass = {cs.CV},
  doi = {10.48550/arXiv.2603.15118},
  url = {https://arxiv.org/abs/2603.15118}
}

@misc{zhang2026paddleocrvl16expandingfrontierdocument,
      title={PaddleOCR-VL-1.6: Expanding the Frontier of Document Parsing with Under-Optimized Region Refinement and Progressive Post-Training}, 
      author={Zelun Zhang and Hongen Liu and Suyin Liang and Yubo Zhang and Yiqing Xiang and Jiaxuan Liu and Ting Sun and Manhui Lin and Yue Zhang and Changda Zhou and Tingquan Gao and Cheng Cui and Yi Liu and Dianhai Yu and Yanjun Ma},
      year={2026},
      eprint={2606.03264},
      archivePrefix={arXiv},
      primaryClass={cs.CV},
      url={https://arxiv.org/abs/2606.03264}, 
}
